%% file: root.tex
\documentclass[letterpaper, 10 pt, conference]{ieeeconf}
\IEEEoverridecommandlockouts
\usepackage{graphics} 
\usepackage{epsfig} 
\usepackage{times} 
\usepackage{amsmath}
\usepackage{amssymb}
\usepackage{graphicx}
\usepackage{tabularx}
\usepackage{caption} 
\usepackage{wrapfig}
\usepackage{romannum}
\usepackage{url}
\usepackage{bm}
\usepackage[table,xcdraw]{xcolor}
\usepackage{tabularray}
\usepackage{cite}
\usepackage[colorlinks=true, linkcolor=red, citecolor=red]{hyperref}
\usepackage{booktabs}
\usepackage{pifont}
\usepackage{bbold}
\usepackage{mathtools}
\usepackage{multirow}
\usepackage{makecell}  
\usepackage{xspace} 
\usepackage{arydshln}
\usepackage{subcaption} 
\input{commands}
\DeclareCaptionFont{mysize}{\fontsize{8}{9.6}\selectfont}

\title{\LARGE \bf
E2-BKI: Evidential Ellipsoidal Bayesian Kernel Inference \\for Uncertainty-aware Gaussian Semantic Mapping
}

\author{Junyoung Kim$^{1}$, Minsik Jeon$^{2}$, Jihong Min$^{1}$, Kiho Kwak$^{1}$, Junwon Seo$^{2}$%
    \thanks{This work was supported by the Agency for Defense Development Grant funded by the Korean Government in 2025.}%
    \thanks{$^{1}$Agency for Defense Development, Daejeon, Republic of Korea. {\tt\footnotesize \{junyoung.kimv, happymin77, kkwak.add\}@gmail.com}}%
    \thanks{$^{2}$Robotics Institute, Carnegie Mellon University, Pittsburgh, PA, USA. {\tt\footnotesize \{minsikj, junwonse\}@andrew.cmu.edu}}%
}

\begin{document}
\maketitle
\begin{abstract}
Semantic mapping aims to construct a 3D semantic representation of the environment, providing essential knowledge for robots operating in complex outdoor settings. While Bayesian Kernel Inference~(BKI) addresses discontinuities of map inference from sparse sensor data, existing semantic mapping methods suffer from various sources of uncertainties in challenging outdoor environments. To address these issues, we propose an uncertainty-aware semantic mapping framework that handles multiple sources of uncertainties, which significantly degrade mapping performance.
Our method estimates uncertainties in semantic predictions using Evidential Deep Learning and incorporates them into BKI for robust semantic inference. It further aggregates noisy observations into coherent Gaussian representations to mitigate the impact of unreliable points, while employing geometry-aligned kernels that adapt to complex scene structures. These Gaussian primitives effectively fuse local geometric and semantic information, enabling robust, uncertainty-aware mapping in complex outdoor scenarios. Comprehensive evaluation across diverse off-road and urban outdoor environments demonstrates consistent improvements in mapping quality, uncertainty calibration, representational flexibility, and robustness, while maintaining real-time efficiency. Our project website: \href{https://e2-bki.github.io/}{\tt\footnotesize https://e2-bki.github.io/}
\end{abstract}
\section{INTRODUCTION}
Semantic mapping constructs a \textit{what-is-where} map by estimating semantic labels at each 3D location, enabling robots to understand and operate in complex environments~\cite{94_kim20133d, 103_valentin2013mesh, 95_SemanticOctree_sengupta2015, 109_paz2020probabilistic, 23_SSMI_asgharivaskasi2023, 26_morilla2023, 106_MapOverconfidence_marques2023}. This process typically operates on sparse and noisy sensor data, along with semantic predictions from neural networks. As sparse observations often lead to discontinuous geometric reconstructions, \textit{continuous mapping} approaches have been proposed to densify sparse regions. A representative approach is Bayesian Kernel Inference (BKI)~\cite{49_BGKOctoMap_doherty2017}, with its semantic extension, S-BKI~\cite{51_S-BKI_gan2020}, which leverages local information to infer the semantics of sparse regions. Through kernel-based spatial propagation, S-BKI addresses the discontinuity issues, enabling spatially consistent semantic mapping.

\begin{figure}[t]
\centering
\includegraphics[width=1.0\linewidth]{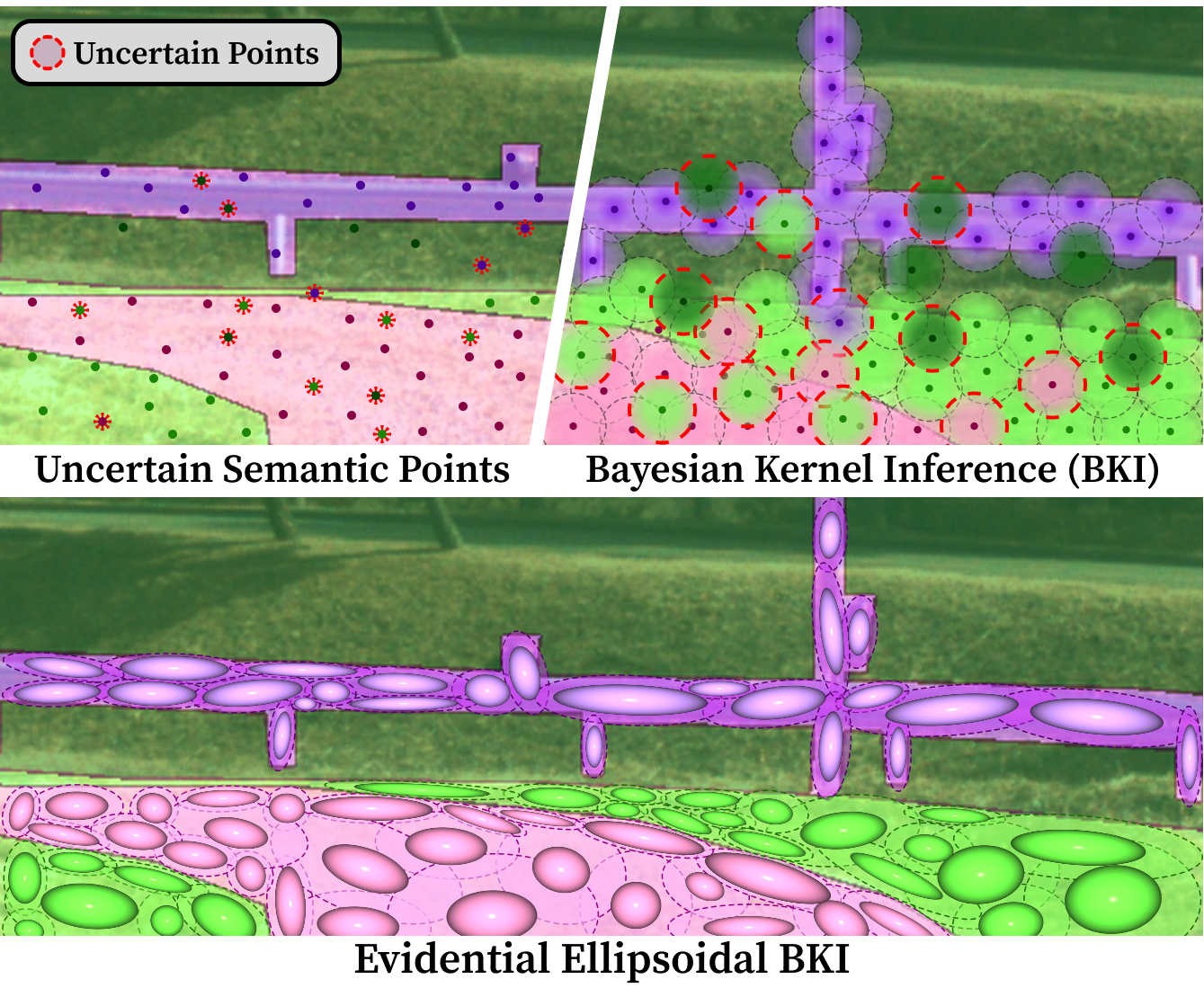}
\vspace{-0.1in}
\caption{\textbf{Evidential Ellipsoidal BKI (E2-BKI)} combines uncertainty-aware processing that prioritizes reliable observations with anisotropic kernels that align with local scene geometry, enabling the construction of accurate and reliable semantic maps from sparse, noisy, and uncertain semantic points.}
\label{fig:concept}
\vspace{-0.25in}
\end{figure}

While S-BKI provides continuity in semantic map inference, it struggles in challenging outdoor conditions due to its inability to account for inherent uncertainties arising from various sources. \textit{Semantic uncertainty} in neural network predictions can undermine mapping performance, as S-BKI treats all observations as equally reliable. \textit{Spatial uncertainty} also arises from the use of static isotropic kernels, which apply uniform influence in all directions and can misalign with anisotropic scene structures such as roads or fences. Moreover, the point-wise processing in BKI fails to account for \textit{observation uncertainty} as it processes points independently, making it vulnerable to noisy and sparse sensor measurements. Although recent works have separately addressed semantic~\cite{EBS_kim2024, DST_kim2024} or spatial uncertainty~\cite{88_SEE-CSOM_deng2023, 47_ConvBKI2_wilson2023}, a unified approach has not been proposed.

To address these limitations, we extend the uncertainty-aware semantic mapping framework~\cite{EBS_kim2024} by introducing \textit{anisotropic} Gaussian primitives as a compact scene representation. Our key insight is that aggregating noisy observations into ellipsoidal primitives enables joint modeling of local geometry, semantics, and uncertainty, while improving robustness to noise. Building on this idea, we construct Gaussian primitives via uncertainty-aware spatial aggregation of neighboring observations, as illustrated in \Fref{fig:concept}. These primitives undergo refinement through merging and pruning, and form the basis of our evidential ellipsoidal BKI formulation with anisotropic kernels that adapt to geometry and incorporate semantic uncertainty. Extensive experiments across both off-road and urban outdoor environments demonstrate superior performance, consistently outperforming existing methods in semantic accuracy and geometric completeness while maintaining real-time efficiency.
In summary, our contributions include:
\begin{itemize}
    \item An uncertainty-aware continuous 3D semantic mapping framework that extends BKI with anisotropic Gaussian primitives, jointly modeling local geometry, semantics, and uncertainty for robust semantic inference.
    \item Evidential Ellipsoidal BKI with geometry-aligned anisotropic kernels that conform to local scene structure and incorporate semantic uncertainty for adaptive fusion.
    \item Comprehensive experiments demonstrating improved accuracy, reliability, and robustness across diverse outdoor environments with real-time performance.
\end{itemize}

\section{RELATED WORK}
\subsection{Continuous Semantic Mapping}
Voxel representation has been widely adopted in semantic mapping for its simplicity and probabilistic compatibility~\cite{94_kim20133d, 95_SemanticOctree_sengupta2015, 23_SSMI_asgharivaskasi2023, 26_morilla2023, 106_MapOverconfidence_marques2023}, where each voxel independently estimates its occupancy and semantic distribution. However, this independence assumption often results in discontinuous maps under sparse sensor data, motivating continuous mapping approaches that incorporate spatial correlations between neighboring voxels. Semantic Bayesian Kernel Inference~(S-BKI)~\cite{51_S-BKI_gan2020} addresses this limitation via kernel-based probabilistic inference~\cite{52_BKI_vega2014} to interpolate semantic information across space, leveraging neighboring observations to infer semantics at unobserved locations.

Despite its effectiveness, S-BKI relies on a static isotropic kernel, which limits its ability to handle both spatial and semantic uncertainty. To overcome these limitations, several extensions have been proposed that modify the kernels. ConvBKI~\cite{47_ConvBKI2_wilson2023} and SEE-CSOM~\cite{88_SEE-CSOM_deng2023} both address spatial uncertainty caused by geometric misalignment. ConvBKI employs learnable class-wise kernels tailored to each class’s geometry, and SEE-CSOM introduces label inconsistency measures to mitigate overinflation at semantic boundaries. Evidential approaches~\cite{EBS_kim2024, DST_kim2024} focus on semantic uncertainty in neural network predictions~\cite{107_guo2017calibration, 83_ModernReliability_de2023}, integrating uncertainty estimates into BKI through adaptive kernels. However, these methods primarily address a single type of uncertainty and process noisy observations independently, making them vulnerable to noise. These limitations motivate the design of intermediate representations that can capture local geometric structure, enable uncertainty-aware semantic fusion, and enhance robustness by aggregating local context.

\subsection{Gaussian Scene Representation}
Recent advances in Gaussian-based mapping provide a promising basis for such intermediate representations due to their capability for local context abstraction and geometric adaptation. Specifically, GMMap~\cite{4_GMMap_li2024} and GIRA~\cite{2_Gira_goel2024} demonstrate the effectiveness of Gaussians for efficient 3D scene reconstruction. However, these methods focus on geometric structure and do not incorporate semantics or uncertainty modeling. In parallel, recent efforts in semantic occupancy prediction have explored Gaussians~\cite{162_GaussianFormer_huang2024} and evidential uncertainty modeling~\cite{264_EvOcc_kalble2025}, but remain constrained by the lack of precise depth information, limiting their applicability to safety-critical scenarios that require accurate geometric reconstruction. Building on these insights, we propose to leverage anisotropic Gaussian primitives that jointly encode geometry, semantics, and uncertainty for robust continuous semantic mapping in complex environments.

\section{Semantic Bayesian Kernel Inference}
\begin{figure*}[t]
\centering
\includegraphics[width=1.0\linewidth]{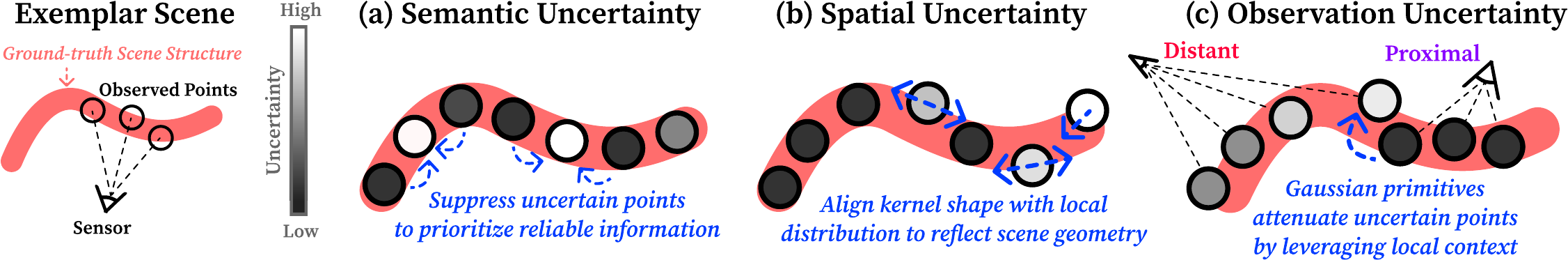}
\vspace{-0.2in}
\caption{\textbf{Key limitations of S-BKI and our solutions.}
The static isotropic kernel in S-BKI does not account for (a) \textit{semantic uncertainty} arising from neural network predictions, (b) \textit{spatial uncertainty} caused by misalignment with local scene geometry, and (c) \textit{observation uncertainty} from distant noisy measurements. Our method addresses these by (a) prioritizing reliable information via uncertainty estimates, (b) adapting kernels to local scene geometry, and (c) leveraging local context through Gaussian primitives.}
\label{fig:cover}
\begin{subfigure}{0pt}\phantomsubcaption\label{fig:cover:a}\end{subfigure}
\begin{subfigure}{0pt}\phantomsubcaption\label{fig:cover:b}\end{subfigure}
\begin{subfigure}{0pt}\phantomsubcaption\label{fig:cover:c}\end{subfigure}
\vspace{-0.35in}
\end{figure*}

We first revisit S-BKI~\cite{51_S-BKI_gan2020}, as a baseline for continuous semantic mapping, and highlight its key limitations that motivate our work. Building upon Evidential Semantic Mapping~(\EBS)~\cite{EBS_kim2024} for semantic uncertainty modeling, we address these limitations through anisotropic Gaussian primitives that enable geometry-aligned inference. For more detailed BKI formulations, we refer readers to~\cite{49_BGKOctoMap_doherty2017, 52_BKI_vega2014}.

\subsection{Semantic Bayesian Kernel Inference}\label{Prelim:BKI}
Let $\dataset$ denote the set of input semantic points, where $\bxn \in \mathbb R^3$ is a 3D coordinate and $\byn \in \{0, 1\}^C$ is a one-hot semantic label over $C$ categories, typically predicted by neural networks~\cite{88_SEE-CSOM_deng2023, 47_ConvBKI2_wilson2023, EBS_kim2024, DST_kim2024}. S-BKI leverages neighboring semantic points to estimate categorical distributions $\hbthm = [\hat{\theta}_m^1, ..., \hat{\theta}_m^C]$ at arbitrary query points $\hbxm$ through kernel inference. The posterior is modeled using the BKI framework~\cite{52_BKI_vega2014}, which extends the standard likelihood by incorporating spatial correlations through kernels. For online robotic applications with sequential semantic points, S-BKI recursively updates Dirichlet posterior parameters that accumulate evidence for each semantic class:
\begin{align}
\label{BKI_derivation}
    p(\hbthm | \hbxm, \mathcal{D}) 
        &\propto p(\mathcal{D} | \hbthm, \hbxm) p(\hbthm | \hbxm) \nonumber \\
        &\propto
            \prod^C_{c=1} (\hat{\theta}_m^c)^{\alpha_0^c + \sum^N_{n=1} k(\hbxm, \bxn)y_n^c - 1},
\end{align}
where $\boldsymbol{\alpha}_0$ represents the Dirichlet prior, and $k$ is the isotropic kernel~\cite{66_SparseKernel_melkumyan2009} that prioritizes nearby observations, with spherical support of radius $\ell$ :
\begin{equation}\label{BKI_5_reparam}
\begin{aligned}
    k(\hbxm, \bxn) &= k'(d, \ell) \\
                &=\underset{d < \ell}{\indicator} \: \Big[ \frac{2 + \cos (2\pi \dol)}{3} (1 - \dol) + \frac{1}{2\pi} \sin (2\pi \dol) \Big] ,
\end{aligned}
\end{equation}
where $\indicator$ is the indicator function, $d = ||\hbxm - \bxn||$ is the distance. Collectively, the posterior parameters $\boldsymbol{\alpha_m}$ incrementally accumulate evidence at each time step $t$:
\begin{equation}
\begin{aligned}\label{BKI_final_update}
    \alpha^c_{m, t} \leftarrow \alpha_{m, t-1}^c + \sum^N_{n=1} k(\hbxm, \bxn) \cdot y^c_n,\quad \alpha^c_{m, 0} = \alpha^c_0,
\end{aligned}
\end{equation} where $\alpha^c_0 \in \mathbb{R}^+$ is an initial hyperparameter for each class. The expectation and variance of $\hbthm$ given the Dirichlet parameters $\boldsymbol{\alpha}_m$ are calculated as:
\begin{equation}
\begin{aligned}\label{BKI_final_variance}
    S_{m} = \sum^C_{c=1} \alpha_{m}^c, \ 
    \mathbb{E}[\hth_{m}^c] = \frac{\alpha_{m}^c}{S_{m}}, \ 
    \mathrm{Var}[\hth_{m}^c] = \frac{\alpha_{m}^c (S_{m} - \alpha_{m}^c)}{S_{m}^2(S_{m} + 1)} .
\end{aligned}
\end{equation}
In this formulation, the semantic map estimate of a query is assigned as $\psi = \arg\max_c \mathbb{E}[\hth_m^c]$, with its variance $\mathrm{Var}[\hth_m^\psi]$ used as a proxy for the uncertainty of the estimate~\cite{51_S-BKI_gan2020, 47_ConvBKI2_wilson2023}.

\subsection{Limitations of S-BKI Framework}\label{Prelim:Limit}
While S-BKI enables continuous semantic mapping, its failures in complex environments can be analyzed in terms of three types of uncertainty. We highlight these three aspects and propose corresponding solutions, as illustrated in \Fref{fig:cover}.

\subsubsection{Semantic Uncertainty}
The update rule \eqref{BKI_final_update} assumes uniform reliability across all observations, neglecting semantic uncertainty from neural network predictions. In visually ambiguous, unfamiliar, or poorly illuminated scenes, the confidence of models varies substantially, and predictions can be inconsistent~\cite{107_guo2017calibration, 83_ModernReliability_de2023}. Therefore, without uncertainty-aware processing, unreliable information is propagated throughout the map with the same weight as reliable predictions, compromising overall mapping quality. The problem intensifies in complex environments where prediction reliability varies widely. To address this limitation, we incorporate semantic uncertainty estimates that enable adaptive weighting of observations, prioritizing confident predictions while attenuating or filtering highly uncertain ones (\Fref{fig:cover:a}).

\subsubsection{Spatial Uncertainty}
The isotropic kernel in \eqref{BKI_5_reparam} applies uniform support across all directions, regardless of local scene geometry. We use the term \textit{spatial uncertainty} to describe the resulting mismatch between this uniform support and the anisotropic nature of real-world structures. When processing elongated structures like roads or fences, spherical kernels propagate semantic information orthogonally to their structural directions, causing geometric misalignment and blurring semantic boundaries, as illustrated in~\Fref{fig:concept}. Moreover, the fixed radius $\ell$ cannot adapt to varying point densities across the scene, leading to excessive smoothing in dense regions and insufficient coverage in sparse areas. Our framework addresses these issues through geometry-aligned anisotropic kernels that adapt their spatial support to local scene structure, steering influence along principal geometric directions while providing adaptive coverage based on local point density (\Fref{fig:cover:b}).

\subsubsection{Observation Uncertainty}
\begin{figure}[t]
\centering
\includegraphics[width=1.0\linewidth]{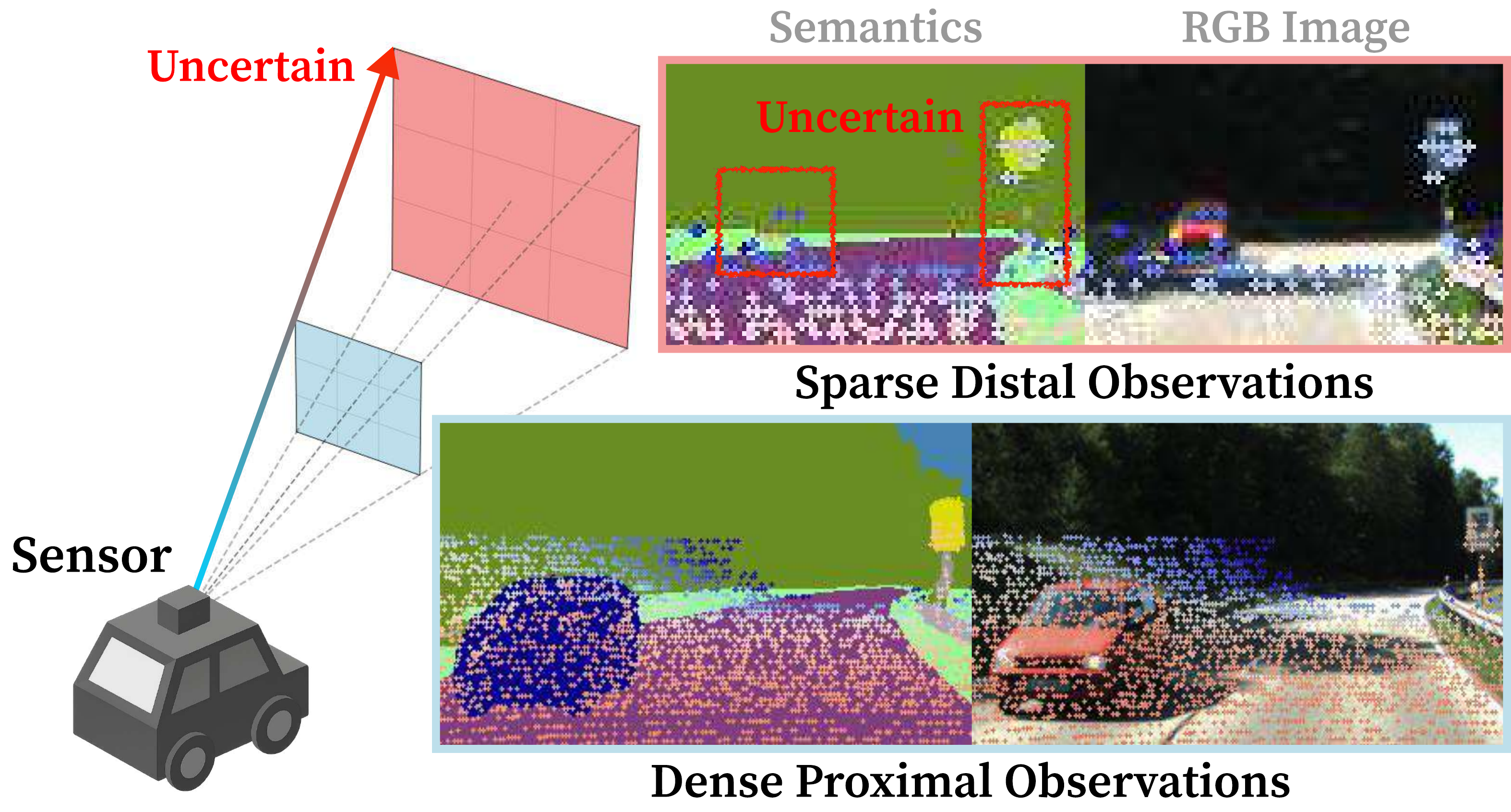}
\vspace{-0.2in}
\caption{Observation uncertainty varies with sensor distance. \textit{Left:} sensor-agnostic schematic showing dense, high-resolution proximal observations and sparse, low-detail distal observations. \textit{Right:} RGB image with semantic labels and LiDAR measurements overlaid; sparser returns at longer ranges lead to higher observation uncertainty.}
\label{fig:uncmodel}
\vspace{-0.25in}
\end{figure}

We use the term \textit{observation uncertainty} to describe the inherent unreliability arising from sensor noise, limited resolution, sparse point coverage, and calibration errors. Although the update rule \eqref{BKI_final_update} incorporates neighboring points for the posterior update, each point is processed independently without leveraging local context. This isolation makes the system susceptible to noise, as individual errors directly affect the semantic map without being corrected by nearby consistent observations. As illustrated in~\Fref{fig:uncmodel}, this susceptibility is further exacerbated with increasing sensor distance, where points become sparser and less informative while calibration errors compound the unreliability of distant observations. These limitations motivate our design of spatially coherent primitives that capture local context, enabling noise reduction via local consistency while providing a more robust foundation for semantic inference (\Fref{fig:cover:c}).

\section{Evidential Ellipsoidal BKI}
\begin{figure*}[t]
\begin{center}
\includegraphics[width=1.0\textwidth]{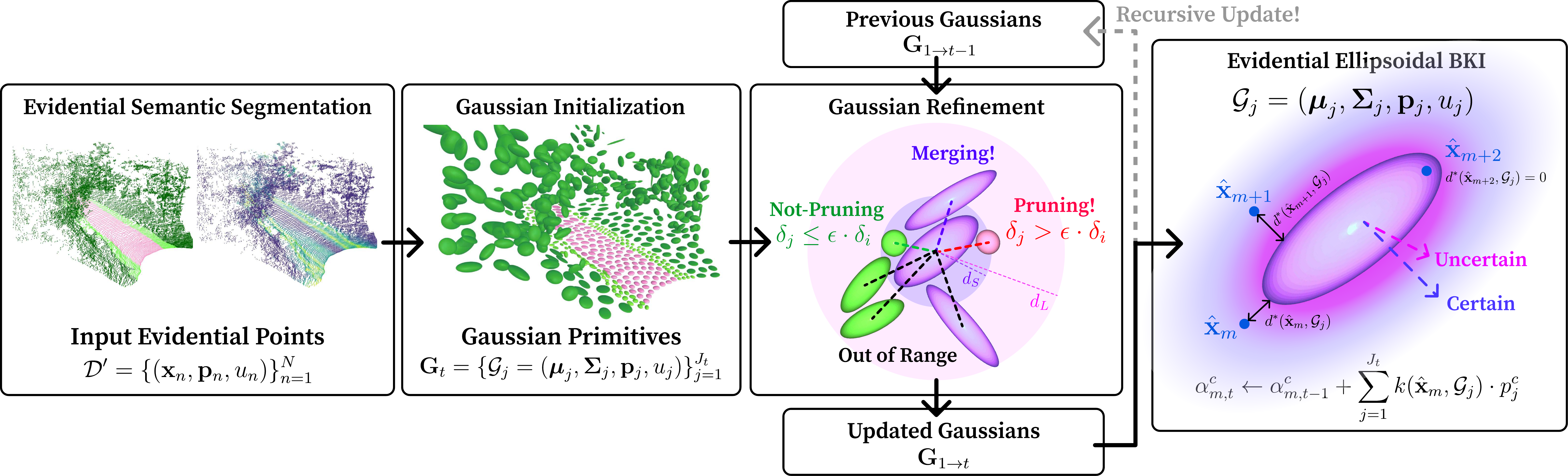}
\end{center}
\caption{\textbf{Overview of semantic mapping with Evidential Ellipsoidal BKI (E2-BKI).} Given evidential points with semantic probability $\mathbf p_n$ and uncertainty $u_n$ (\Sref{Method:EDL}), our method operates through three key stages: (\Sref{Method:Init}) Gaussian Initialization aggregates evidential points into anisotropic Gaussian primitives encoding local geometry and semantics; (\Sref{Method:Refine}) Gaussian Refinement merges spatially coherent primitives and prunes unreliable primitives; and (Sections~\ref{Method:Kernel} and \ref{Method:EEBKI}) Evidential Ellipsoidal BKI performs uncertainty-aware semantic mapping using Gaussian primitives.}
\label{fig:framework}
\vspace{-0.25in}
\end{figure*}

To overcome the limitations identified in \Sref{Prelim:Limit}, we propose an uncertainty-aware mapping framework that simultaneously addresses semantic, spatial, and observation uncertainty through anisotropic Gaussian primitives. An overview of the full pipeline is illustrated in~\Fref{fig:framework}.

\subsection{Evidential Semantic Segmentation}\label{Method:EDL}
To quantify uncertainties of semantic predictions, we adopt Evidential Deep Learning~(EDL)~\cite{10_EDL_sensoy2018} to extend one-hot semantic predictions $\byn$ into semantic probability $\bpn = [p_n^1, \ldots, p_n^C]$ with uncertainty estimates $u_n$, extending the input $\mathcal{D}$ to evidential points $\mathcal{D'} = \{ (\bxn, \bpn, u_n) \}^N_{n=1}$. This formulation follows \EBS~\cite{EBS_kim2024}, which presents EDL-based uncertainty estimates of semantic prediction and its integration into uncertainty-aware semantic mapping. Based on Dempster-Shafer Theory of evidence~(DST)~\cite{129_DST_yager2008}, EDL estimates per-class evidence in a single forward pass and converts it into class probability $\bpn$ and corresponding uncertainty $u_n$. These uncertainty estimates reflect the model’s confidence in its predictions, which helps reduce the impact of uncertain predictions during fusion (see~\cite{EBS_kim2024} for details). The evidential points are then aggregated into Gaussian primitives through uncertainty-aware fusion to construct robust semantic representations.

\subsection{Gaussian Initialization}\label{Method:Init}
While evidential points contain valuable semantic information with uncertainty, individual semantic predictions are unstable and fail to capture spatial context. This limitation motivates the aggregation of evidential points into structured representations that capture local context and reduce sensitivity to noise. Specifically, we adopt 3D Gaussian primitives for their natural ability to capture spatial anisotropy in a compact and mathematically tractable form.

To this end, we begin by clustering the evidential points into $J$ clusters $\{ \mathcal{C}_j \}_{j=1}^{J}$. We use the K-Means++ algorithm~\cite{267_KMenasPP_arthur2006} as a stable default choice, although other clustering strategies are also compatible with our framework.
Each cluster $\mathcal{C}_{j}$ is then abstracted as a \textit{Gaussian primitive} $\mathcal G_j = (\bmu_j, \bSigma_j, \mathbf p_j, u_j)$, where the local geometry is modeled by mean $\bmu_j$ and covariance $\bSigma_j$. The semantic probability $\mathbf p_j$ and uncertainty $u_j$ are derived by aggregating the semantic predictions $(\bpn, u_n)$ from all points in $\mathcal{C}_j$ through DST-based combination rules~\cite{72_TMCJournal_han2022}. Specifically, we convert each probability $\mathbf p_n$ into belief mass $\mathbf b_n$ with $b_n^c = p_n^c - u_n / C $, and apply the combination rule for combining two belief masses:
\begin{equation}\label{dst_fusion}
    b^c = \frac{1}{1-\eta} (b_1^c b_2^c + b_1^c u_2 + b_2^c u_1),\ \ u = \frac{1}{1-\eta} u_1 u_2,
\end{equation}
where $C$ is the number of semantic classes, and $\eta = \sum_{x \ne y} b_1^x b_2^y$ is a measure of the conflict between two belief masses. We iteratively apply this pairwise combination rule to fuse all $|\mathcal{C}_j|$ belief masses within the cluster, progressively accumulating evidence. Then, the merged belief mass $\mathbf b_j$ is converted back into the probability via $p_j^c = b_j^c + u_j / C$. This uncertainty-aware fusion effectively resolves semantic conflicts within each cluster and yields robust semantics with reliable uncertainty estimates (for more details, see~\cite{DST_kim2024}).

\subsection{Gaussian Refinement}\label{Method:Refine} 
\subsubsection{Merging}
As BKI incrementally accumulates evidence over time (\Sref{Prelim:BKI}), Gaussian primitives similarly accumulate through iterative initialization. This temporal accumulation can cause redundancy from overlapping regions, while some primitives may consist of insufficient points for stable geometric estimation.
To address these issues, we merge nearby primitives with the same semantic label. Geometric components $(\bmu_j, \bSigma_j)$ are merged by combining their statistical moments, while semantic components $(\mathbf p_j, u_j)$ are fused through DST-based combination rules as in~\eqref{dst_fusion}. This merging stabilizes both geometric and semantic representations and reduces computational overhead by decreasing the total number of primitives.

\subsubsection{Pruning} Additionally, accumulated primitives may exhibit semantic inconsistencies within local regions. These inconsistencies primarily arise from uncertain distant observations, which are both spatially sparse and semantically unreliable~(\Fref{fig:uncmodel}). Inspired by human visual processing that maintains initial impressions until contradicted by more reliable evidence, we adopt a relative pruning strategy. We assess the reliability of each primitive $\mathcal G_j$ using its sensor distance $\delta_j$, computed as the average distance from the sensor to the points aggregated in that primitive. Primitives are pruned only when a neighboring primitive $\mathcal G_i$ with significantly lower $\delta_i$ has conflicting semantics:
\begin{equation}\label{prune_cond}
    \argmax\ \mathbf{p}_i \ne \argmax\ \mathbf{p}_j\quad \text{and} \quad \delta_j > \epsilon \cdot \delta_i,
\end{equation}
where $\epsilon$ controls the pruning sensitivity. This strategy selectively removes unreliable observations only when contradicted by more reliable ones, preserving uncertain yet potentially informative observations.

\subsection{Evidential Ellipsoidal Kernel}\label{Method:Kernel}
To perform BKI with our Gaussian primitives, we design an evidential ellipsoidal kernel that exploits both the anisotropic geometry and semantic uncertainty of each primitive. The kernel adapts its spatial support to primitive geometry, while incorporating uncertainty estimates to filter unreliable primitives and prioritize reliable ones.

For geometry-aligned kernel support, we compute the distance $d^\ast(\hbxm, \mathcal G_j)$ by representing each primitive as an anisotropic ellipsoid and finding the closest point $\mathbf v^{\ast}$ on the ellipsoid surface to query point $\hbxm$ through the following minimization over surface points $\mathbf v$:
\begin{equation}
d^{\ast} = \min_{\mathbf{v} \in \mathbb{R}^3} \|\hbxm - \mathbf{v}\|^2 \ \text{s.t.} \ (\mathbf{v} - \boldsymbol{\mu}_{j})^\top {\boldsymbol{\Sigma}_{j}}^{-1} (\mathbf{v} - \boldsymbol{\mu}_{j}) = \tau ,
\end{equation}
where $\tau$ controls the ellipsoid size and the resulting minimum $d^\ast(\hbxm, \mathcal G_j) := d^{\ast}$ used in our kernel. For queries inside the ellipsoid, we set $d^\ast(\hbxm, \mathcal G_j) := 0$. This formulation enables kernels to conform to elongated structures and propagate semantic information along principal geometric directions, addressing the spatial uncertainty of isotropic kernels.

Using this anisotropic distance, our kernel incorporates semantic uncertainty $u_j$ to adaptively modulate kernel support and filter unreliable primitives. We extend the uncertainty-aware kernel function of the evidential semantic mapping framework (\EBS~\cite{EBS_kim2024}) with a geometry-aligned formulation:
\begin{equation}
\begin{aligned}\label{E2BKI_kernel}
  \tilde k(\hbxm, \mathcal G_j) =
      &\begin{cases}
          k'(d^\ast(\hbxm, \mathcal G_j), \ell \cdot \beta e ^{1 - u_j}) &\text{if } u_j \le \Uthres \\
          0 &\text{if } u_j > \Uthres ,
      \end{cases}
\end{aligned}
\end{equation}
where $\beta$ controls uncertainty sensitivity and $\Uthres$ is dynamically set to exclude the most uncertain $\tilde{u}$ percentile of primitives. This formulation ensures that highly uncertain observations are filtered out while smoothly modulating spatial influence based on semantic uncertainty, enabling robust semantic inference (for more details, see \cite{EBS_kim2024}). When multiple primitives overlap, their kernel-weighted evidence is accumulated in $\balpha_m$, so primitives with lower uncertainty naturally dominate the posterior semantics while uncertain primitives contribute only weak evidence.

\subsection{Evidential Ellipsoidal BKI}\label{Method:EEBKI}
We integrate the evidential ellipsoidal kernel into the BKI framework, enabling continuous semantic mapping with spatial and semantic adaptability. Building upon \EBS~\cite{EBS_kim2024}, our method operates on Gaussian primitives with probabilistic semantics $\mathbf p_j$ rather than with discrete labels:
\begin{equation}
\begin{aligned}\label{E2BKI_final_update}
  \alpha^c_{m,t} \leftarrow \alpha_{m,t-1}^c + \sum^J_{j=1} \tilde k(\hbxm, \mathcal G_j) \cdot p^c_j.
\end{aligned}
\end{equation}
This formulation addresses all three limitations in~\Sref{Prelim:Limit}: \textit{semantic uncertainty} is mitigated through evidential fusion, \textit{spatial uncertainty} is resolved via geometry-aligned kernels, and \textit{observation uncertainty} is reduced by leveraging local context through Gaussian primitives and uncertainty-aware pruning. These components collectively yield balanced semantic estimates with reduced noise sensitivity, enabling robust semantic inference in complex outdoor environments.

\section{EXPERIMENTS}
\begin{table}[b]
\vspace{-0.15in}
\caption{Hyperparameters for mapping frameworks.}
\vspace{-0.05in}
\centering
\renewcommand{\arraystretch}{1.0}
\resizebox{1.0\linewidth}{!}{%
    \begin{tabular}{c cccccccc}
        \toprule
        \multirow{2}{*}{Symbol} & $\alpha_0^c$ & $\ell$ & $\beta$ & $\Tilde{u}$ & $\epsilon$ & $d_L$ & $d_S$ \\
        & (\ref{Prelim:BKI}) & (\ref{Method:Kernel}) & (\ref{Method:Kernel}) & (\ref{Method:Kernel}) 
           & (\ref{Method:Refine}) & (\ref{Method:Refine}) & (\ref{Method:Refine}) \\
        \midrule
        Value & $0.001$      & $0.2\textrm{m}$ & $0.75$ & $10\%$ & $2.5$   & $5\ell$ & $\ell$ \\
        \bottomrule
    \end{tabular}
}
\label{table:hyperparams}
\end{table}

We validate our evidential ellipsoidal BKI framework through comprehensive experiments designed to answer four key research questions: (1) Does our unified uncertainty handling approach improve mapping quality across large-scale off-road and urban environments in terms of both semantic accuracy and uncertainty calibration? (2) How robust is our framework to challenging conditions such as sparse input data and unreliable semantic predictions? (3) Does our Gaussian primitive representation provide versatility across different map formats? (4) What is the individual contribution of each component, and does the framework maintain real-time efficiency?

\subsection{Datasets}
\subsubsection{Off-road Environments}
Our evaluation includes two off-road datasets. The first, \RELLIS~\cite{122_RELLIS_jiang2021}, provides RGB images and OS1-64 LiDAR scans with 2D semantic annotations and accurate robot poses. We conduct a five-fold evaluation by holding out each sequence once. The second is an extended version of \OFFROAD~\cite{EBS_kim2024}, offering broader spatial coverage and collected using a platform equipped with an OS1-128 LiDAR and RGB camera. It includes manually annotated RGB images and employs geographically disjoint train-test splits. Collected under diverse seasonal, lighting, and terrain conditions, \OFFROAD presents significant challenges for reliable semantic mapping.

\subsubsection{Urban Environments} For comprehensive evaluation, we also include the \KITTI~\cite{265_KITTI360_liao2022} dataset, which spans 73.7 km of urban driving. We use front-view RGB images and HDL-64E LiDAR scans, along with 2D semantic annotations and GPS/IMU-based localization. We use sequences 00, 02-06 for training; 07, 09 for validation; and 10 for testing.

\subsection{Experimental Setup}
\subsubsection{Implementation Details}
We employ LRASPP~\cite{266_LRASPP_howard2019} with EDL~\cite{10_EDL_sensoy2018} for 2D semantic segmentation with semantic uncertainty estimation. The network is trained for 15 epochs on off-road datasets and 10 epochs on urban datasets. Other training details follow the configuration in \EBS~\cite{EBS_kim2024}. The 2D semantic predictions are projected onto 3D point clouds to construct the input $\mathcal{D}'$.

For 3D mapping, we perform a grid search to select the hyperparameters. The chosen values are listed in Table~\ref{table:hyperparams}, and we note that E2-BKI is not sensitive to these values. Gaussian initialization first partitions points by semantic class, then applies K-means++ clustering within each partition. We use 256 clusters for \RELLIS and 1024 for other datasets to account for variations in point density. During refinement, we first check for semantic consistency within $d_L$ neighborhoods: if all neighboring primitives are semantically consistent, those within $d_S$ are merged. When semantic conflicts exist within $d_L$ and condition \eqref{prune_cond} is satisfied, the more distant primitive is pruned. The threshold $\tau$ is chosen to enclose $10\%$ of the probability mass of each Gaussian. 

\subsubsection{Evaluation Metrics}
For fair comparison, all methods are evaluated on a $0.2\textrm{ m}$ voxelized grid. We adopt metrics established in \EBS~\cite{EBS_kim2024}, including per-class \IoU and \mIoU for semantic accuracy with extending the Accuracy to $\mathrm{Acc} = \sum_{c=1}^{C} \tpk / \mathbf{Q}$ for joint geometric and semantic evaluation, where $\tpk$ denotes true positives for class $c$ and $\mathbf{Q}$ is the total number of queries. For uncertainty calibration, we employ Brier Score~(\BS$\downarrow$) with normalized $\mathrm{Var}[\hth_m^\psi]$.

\subsubsection{Comparison Methods}
We include representative baselines where S-CSM~\cite{51_S-BKI_gan2020} serves as the discrete mapping baseline, and S-BKI~\cite{51_S-BKI_gan2020}, ConvBKI~\cite{47_ConvBKI2_wilson2023}, SEE-CSOM~\cite{88_SEE-CSOM_deng2023}, and \EBS~\cite{EBS_kim2024} as continuous mapping approaches. SEE-CSOM~\cite{88_SEE-CSOM_deng2023} addresses spatial uncertainty by incorporating label inconsistency measures to mitigate overinflation at the semantic boundary. ConvBKI~\cite{47_ConvBKI2_wilson2023} learns class-specific kernels, necessitating dataset-specific training. Consequently, we train separate models for each dataset.

\begin{figure*}[t!]
\begin{center}
\includegraphics[width=1.0\textwidth]{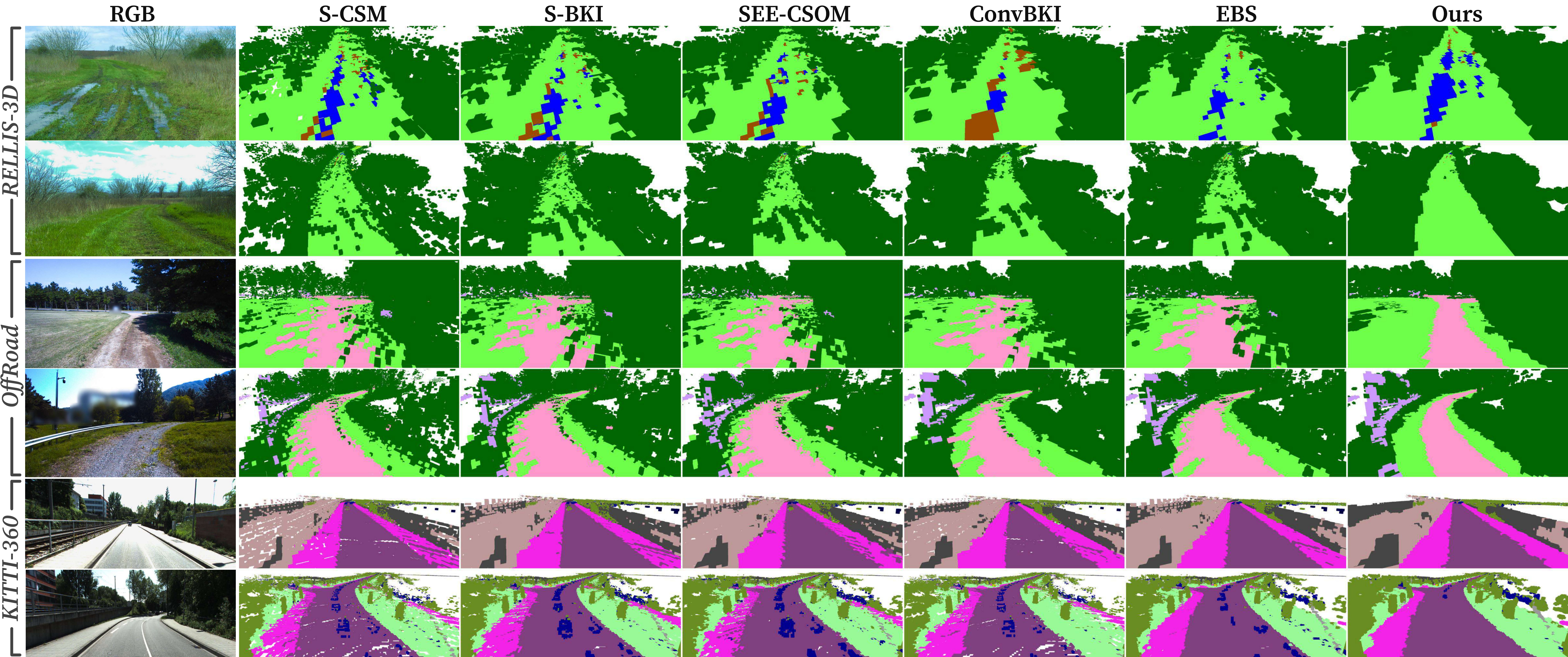}
\end{center}
\vspace{-0.15in}
\caption{Qualitative comparison of semantic mapping results on \RELLIS, \OFFROAD, and \KITTI. Compared to baselines, our method produces more accurate and visually consistent semantic reconstructions across diverse scenes. The color scheme for semantic classes follows \Tref{tab:offroad_quant} and \Tref{tab:urban_quant}.}
\label{fig:mainqualitative}
\vspace{-0.15in}
\end{figure*}

\begin{table}[t!]
\centering
\renewcommand {\arraystretch}{1.3}
\caption{Quantitative results on off-road environments (\RELLIS and \OFFROAD). Semantic classes not present in the dataset are excluded from \mIoU and \BS and are indicated by a dash (-).  Our method consistently outperforms or matches prior continuous semantic mapping approaches.
}
\label{tab:offroad_quant}
\Large{
\resizebox{1.0\linewidth}{!}{
    \begin{tabular}{c | l | ccccccc | ccc}
    \toprule
    \multicolumn{2}{c}{} & \multicolumn{7}{c}{Per-class IoU (\%)} & \multicolumn{3}{c}{} \\
    \cmidrule(lr){3-9}
    \textbf{Dataset}
    & \textbf{Method}
    & \rotatebox{90}{\semcolor[puddle]     \hspace{0pt}puddle}
    & \rotatebox{90}{\semcolor[object]     \hspace{0pt}object}
    & \rotatebox{90}{\semcolor[paved]      \hspace{0pt}paved}
    & \rotatebox{90}{\semcolor[unpaved]    \hspace{0pt}unpaved}
    & \rotatebox{90}{\semcolor[dirt]       \hspace{0pt}dirt}
    & \rotatebox{90}{\semcolor[grass]      \hspace{0pt}grass}
    & \rotatebox{90}{\semcolor[vegetation] \hspace{0pt}vegetation}
    & \rotatebox{90}{\mIoU [\%]} 
    & \rotatebox{90}{\Acc [\%]}  
    & \rotatebox{90}{\BS$\downarrow$ [\%]}   
    \\
    \midrule \midrule
\multirow{6}{*}{\shortstack{\textbf{\textit{RELLIS}} \\ \textbf{\textit{3D}}}} 
    & \textit{S-CSM~\cite{51_S-BKI_gan2020}}&
    {37.7} & {8.1} & {45.5} & - & \first{19.5} & {74.5} & {77.0} & {43.7} & {67.7} & {15.0} \\
    & \textit{S-BKI~\cite{51_S-BKI_gan2020}}&
    {35.9} & {8.3} & {49.4} & - & \secnd{19.0} & {73.8} & {76.6} & {43.8} & {78.9} & {16.3} \\
    & \textit{SEE-CSOM~\cite{88_SEE-CSOM_deng2023}} & 
    {36.5} & {10.0} & {51.5} & - & {18.6} & {74.0} & {77.2} & {44.6} & {80.0} & {14.1} \\
    & \textit{ConvBKI~\cite{47_ConvBKI2_wilson2023}} &
    {35.1} & \first{11.4} & \secnd{56.0} & - & {11.2} & \secnd{75.2} & {76.8} & {44.3} & {80.7} & {15.2} \\
    & \textit{\EBS~\cite{EBS_kim2024}} &
    \secnd{39.8} & {8.2} & {55.0} & - & {15.6} & {74.4} & {77.6} & \secnd{45.1} & \secnd{80.9} & \secnd{13.2} \\
    & \textit{Ours} &
    \first{44.8} & \secnd{10.7} & \first{57.9} & - & {16.1} & \first{75.3} & \first{79.2} & \first{47.3} & \first{83.5} & \first {13.0} \\
\hline
\multirow{6}{*}{\shortstack{\textbf{\OFFROAD}}}
    & \textit{S-CSM~\cite{51_S-BKI_gan2020}}&
    - & {45.9} & - & {80.9} & - & {61.9} & {89.8} & {69.6} & {58.5} & {9.2}\\
    & \textit{S-BKI~\cite{51_S-BKI_gan2020}}&
    - & {43.6} & - & {80.8} & - & {62.7} & {89.8} & {69.2} & {78.5} & {14.4}\\
    & \textit{SEE-CSOM~\cite{88_SEE-CSOM_deng2023}} &
    - & {42.9} & - & {80.2} & - & {62.2} & {89.9} & {68.8} & {80.3} & {8.7}\\
    & \textit{ConvBKI~\cite{47_ConvBKI2_wilson2023}} &
    - & \secnd{46.7} & - & {82.0} & - & \secnd{65.4} & {90.5} & {71.1} & \secnd{84.3} & {8.4}\\
    & \textit{\EBS~\cite{EBS_kim2024}} &
    - & \first{48.1} & - & \secnd{83.4} & - & {65.2} & \secnd{91.1} & \first{71.9} & {82.6} & \first{7.0}\\
    & \textit{Ours} &
    - & {43.9} & - & \first{84.1} & - & \first{66.5} & \first{91.4} & \secnd{71.5} & \first{87.7} & \secnd{7.4}\\
    
    \bottomrule
    \end{tabular}
    }
}
\vspace{-0.1in}
\end{table}

\subsection{Evaluations on Semantic Mapping Performance} \subsubsection{Semantic Accuracy}
Our evidential ellipsoidal BKI framework consistently outperforms existing methods across diverse environments (\Tref{tab:offroad_quant}, \Tref{tab:urban_quant}). Across all datasets, our approach achieves the highest \Acc, indicating superior performance in both geometric completeness and semantic accuracy. These improvements stem from geometry-aligned ellipsoidal kernels and primitive-based processing that mitigate spatial uncertainty, while comprehensive uncertainty handling ensures robust semantic representations by prioritizing reliable predictions.
Our method also achieves the best \mIoU on \RELLIS and \KITTI, with competitive results on \OFFROAD. The slight \mIoU reduction on \OFFROAD relative to \EBS reflects an inherent trade-off: improving geometric completeness increases the number of difficult queries in sparse regions, making semantic classification more challenging. Nevertheless, the overall \Acc gains outweigh this \mIoU difference in practical robotic applications where both geometry and semantics matter.

\begin{table}[t!]
\centering
\renewcommand {\arraystretch}{1.3}
\caption{Quantitative results on \KITTI. Our method shows superior performance over prior continuous semantic mapping approaches.}
\label{tab:urban_quant}
\Large{
\resizebox{1.0\linewidth}{!}{
    \begin{tabular}{l | ccccccccccc | ccc}
    \toprule
    \multicolumn{1}{c}{} & \multicolumn{11}{c}{Per-class IoU (\%)} & \multicolumn{3}{c}{} \\
    \cmidrule(lr){2-12}
    \textbf{Method}
    & \rotatebox{90}{\semcolor[road]     \hspace{0pt}road}
    & \rotatebox{90}{\semcolor[sidewalk]     \hspace{0pt}sidewalk}
    & \rotatebox{90}{\semcolor[building]      \hspace{0pt}building}
    & \rotatebox{90}{\semcolor[fence]    \hspace{0pt}fence}
    & \rotatebox{90}{\semcolor[pole]       \hspace{0pt}pole}
    & \rotatebox{90}{\semcolor[sign]      \hspace{0pt}sign/light}
    & \rotatebox{90}{\semcolor[kittivegetation] \hspace{0pt}vegetation}
    & \rotatebox{90}{\semcolor[terrain]       \hspace{0pt}terrain}
    & \rotatebox{90}{\semcolor[person]      \hspace{0pt}person}
    & \rotatebox{90}{\semcolor[car] \hspace{0pt}car}
    & \rotatebox{90}{\semcolor[truck] \hspace{0pt}truck}
    & \rotatebox{90}{\mIoU [\%]} 
    & \rotatebox{90}{\Acc [\%]}  
    & \rotatebox{90}{\BS$\downarrow$ [\%]}   
    \\
    \midrule \midrule
    \textit{S-CSM~\cite{51_S-BKI_gan2020}} &
        {88.8} & {61.5} & {68.3} & {47.1} & \first{15.8} & {11.4} & {68.9} & {59.0} & {8.0} & \first{67.1} & {23.9} & {47.3} & {59.5} & {14.7}\\
    \textit{S-BKI~\cite{51_S-BKI_gan2020}} & 
        {88.3} & {61.3} & {66.7} & {47.7} & \secnd{15.0} & {10.8} & {68.6} & {60.4} & {8.2} & {65.5} & {25.9} & {47.1} & {74.9} & {17.4}\\
    \textit{SEE-CSOM~\cite{88_SEE-CSOM_deng2023}} &
        {88.5} & {61.8} & {66.8} & {47.8} & {14.2} & {9.0} & {68.5} & {60.2} & {8.0} & {65.6} & {25.0} & {46.9} & {76.1} & {13.4}\\
    \textit{ConvBKI~\cite{47_ConvBKI2_wilson2023}} &
        {87.6} & {64.1} & \first{69.4} & {48.6} & {13.2} & \secnd{13.3} & \secnd{69.9} & \secnd{63.6} & {8.7} & \secnd{66.9} & {26.6} & {48.3} & {68.0} & {13.5}\\
    \textit{\EBS~\cite{EBS_kim2024}} & 
        \first{90.5} & \first{67.1} & \secnd{69.3} & \secnd{50.2} & {14.0} & {11.9} & \first{70.4} & {63.4} & \secnd{9.2} & {65.6} & \secnd{27.2} & \secnd{49.0} & \secnd{77.8} & \secnd{12.5}\\
    \textit{Ours} & 
        \secnd{90.1} & \secnd{67.0} & {68.7} & \first{50.8} & {14.3} & \first{14.2} & \secnd{69.9} & \first{63.7} & \first{9.7} & {64.9} & \first{30.2} & \first{49.4} & \first{80.0} & \first{12.1} \\  
    \bottomrule
    \end{tabular}
    }
}
\vspace{-0.3in}
\end{table}

These quantitative improvements translate into clear visual advantages across all environments, as shown in \Fref{fig:mainqualitative}. On \RELLIS, we preserve puddle boundaries and produce clean traversable paths by suppressing uncertain predictions, while baselines blur edges and yield incorrect labels. In \OFFROAD, the unpaved road remains clearly separated from grass with noise effectively suppressed, whereas baselines exhibit small isolated errors near boundaries. For urban \KITTI, road geometry remains continuous and free from noisy artifacts, owing to our comprehensive handling of local geometric structures and uncertainties.

\subsubsection{Uncertainty Calibration}
\begin{figure*}[!t]
    \centering
    \includegraphics[width=1.0\linewidth]{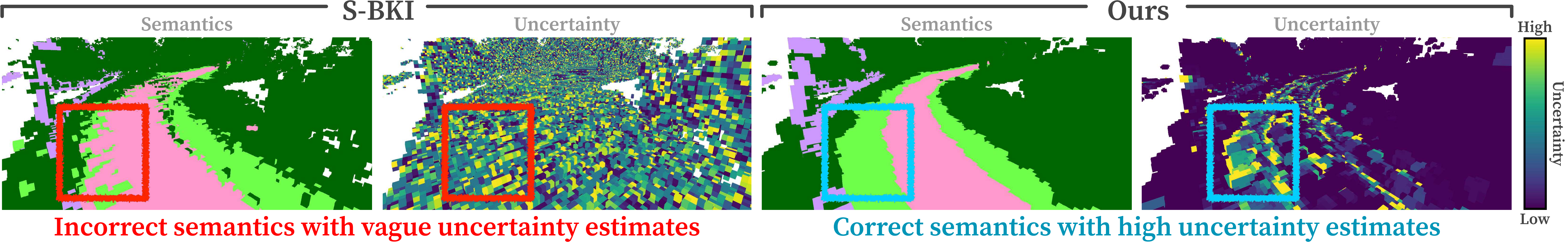}
    \caption{Semantic and uncertainty map (the third row of \Fref{fig:mainqualitative}). While S-BKI produces incorrect semantics with uncalibrated uncertainty estimates, E2-BKI recovers the correct semantic labels and explicitly flags these challenging regions with higher uncertainty.}
    \label{fig:semvar}
\vspace{-0.2in}
\end{figure*}

E2-BKI achieves well-calibrated uncertainty estimates across all datasets, achieving the best \BS on \RELLIS and \KITTI and competitive performance on \OFFROAD. \Fref{fig:semvar} illustrates how our uncertainty-aware processing addresses unreliable semantic predictions. In the boxed region, S-BKI produces incorrect semantics with weakly informative uncertainty, whereas our framework generates more reliable semantic assignments and preserves higher uncertainty in areas where the evidence is insufficient. Consequently, the predicted uncertainty better reflects the difficulty of each prediction and clearly identifies high-risk regions for downstream modules, enabling accurate semantic maps together with well-calibrated and trustworthy uncertainty information across diverse environments.

\begin{figure}[t]
\begin{minipage}{0.48\textwidth}
    \centering
    \includegraphics[width=\linewidth]{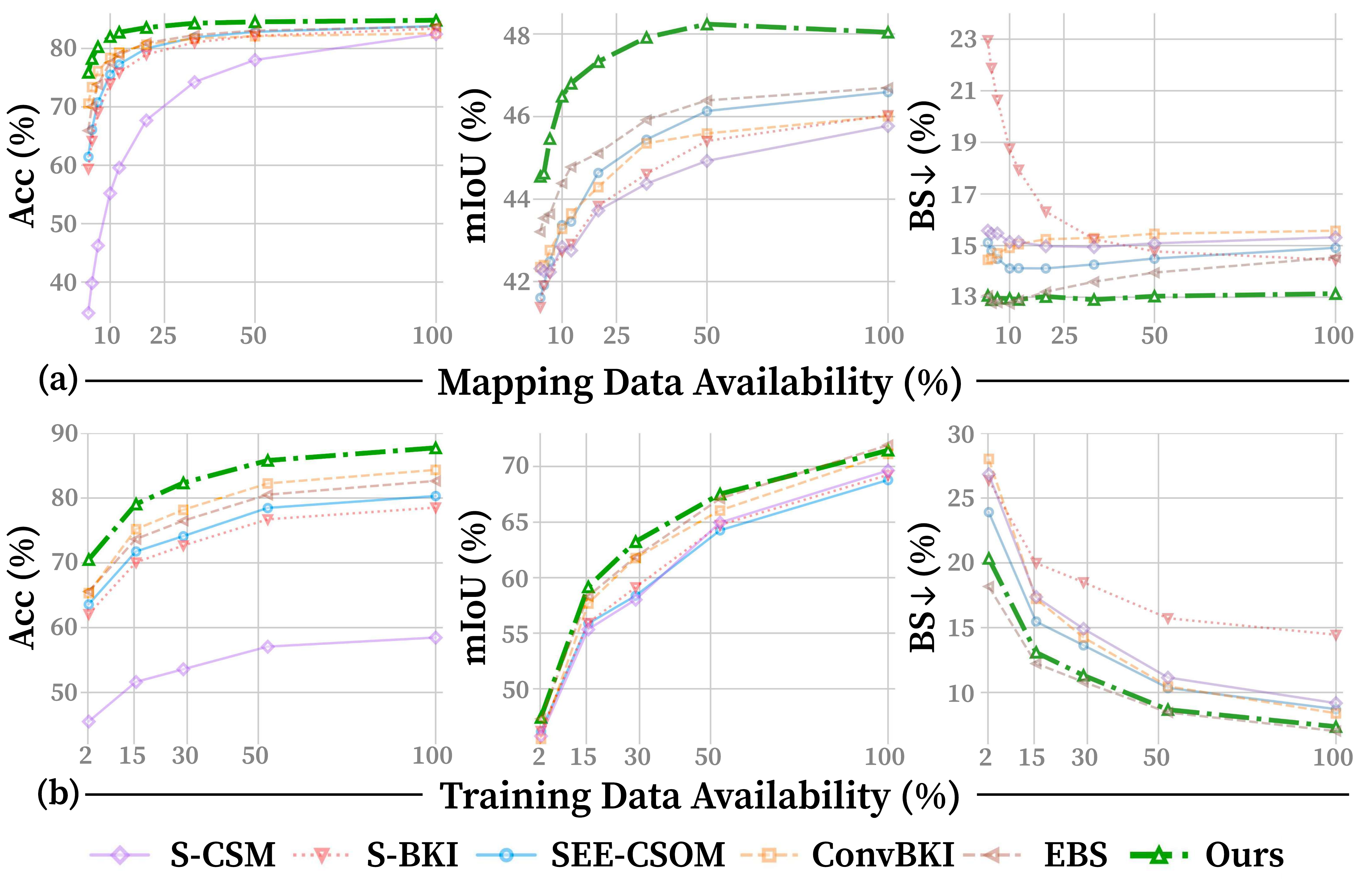}
    \caption{Robustness evaluation under data limitations. (a) Performance under varying input sparsity on \RELLIS. (b) Performance with semantic networks trained on limited data of \OFFROAD. E2-BKI maintains the most robust performance under uncertainties arising from data limitations.}
    \label{fig:robust}
    \begin{subfigure}{0pt}\phantomsubcaption\label{fig:robust:a}\end{subfigure}
    \begin{subfigure}{0pt}\phantomsubcaption\label{fig:robust:b}\end{subfigure}
\end{minipage}
\vspace{-0.4in}
\end{figure}

\subsection{Evaluations on Robustness and Versatility}\label{EXP:RobVer}
\subsubsection{Robustness under Input Sparsity}
To evaluate robustness under sparse sensor data, we reduce the mapping input by using only a subset of available frames while maintaining the same evaluation protocol. \Fref{fig:robust:a} illustrates the robustness of our method to sparse input on \RELLIS, consistently outperforming baselines as input availability decreases to $4\%$. This resilience stems from Gaussian primitives that aggregate sparse observations into coherent structures, capturing local continuity despite missing observations. In contrast, point-based methods do not leverage local context and degrade significantly under sparse conditions.

\subsubsection{Robustness to Unreliable Semantic Predictions}
To evaluate the robustness of mapping performance under degraded semantic predictions, we train semantic segmentation networks on progressively smaller training datasets. \Fref{fig:robust:b} presents results on \OFFROAD with training data reduced to $2\%$. While all methods experience performance degradation, ours consistently achieves the highest \Acc across all settings. This robustness stems from our uncertainty-aware processing, which mitigates the impact of unreliable predictions.

\subsubsection{Scene Representation Versatility}
Our Gaussian primitives offer greater flexibility and versatility compared to traditional voxel-based representations. For instance, they can be directly projected onto 2D planes to form BEV semantic maps. \Tref{tab:bev} demonstrates that our E2-BKI framework effectively transfers to this different map format, achieving consistent improvements across all datasets in BEV projection mode. Beyond BEV mapping, our compact Gaussian representation enables continuous evaluation at arbitrary query points through direct semantic inference. This is achieved by applying our evidential ellipsoidal BKI directly to nearby primitives at any 3D location, eliminating pre-defined voxel grids. Since this continuous mode avoids discretization artifacts by directly querying primitives, further performance improvements are observed across all metrics in \Tref{tab:bev}. This flexibility supports diverse mapping applications, including multi-resolution structures, adaptive sampling strategies, and custom query patterns, highlighting the practical advantages of our primitive-based framework.

\begin{table}[t!]
\centering
\renewcommand{\arraystretch}{1.2}
\caption{
 Results on BEV semantic mapping. Our method consistently improves performance in both BEV projection and continuous evaluation modes, with the latter achieving the best performance via direct inference at ground-truth query points without discretization.
}
\label{tab:bev}
\Large{
\resizebox{1.0\linewidth}{!}{%
    \begin{tabular}{l ccc | ccc | ccc}
        \toprule
        & 
        \multicolumn{3}{c}{\textbf{\RELLIS}} & 
        \multicolumn{3}{c}{\textbf{\OFFROAD}} &
        \multicolumn{3}{c}{\textbf{\KITTI}} \\
        \cmidrule(lr){2-4} \cmidrule(lr){5-7} \cmidrule(lr){8-10}
        {\textbf{Method}} & \mIoU & \Acc & \BS$\downarrow$ & \mIoU & \Acc & \BS$\downarrow$ & \mIoU & \Acc & \BS$\downarrow$ \\
        \midrule \midrule

        \textit{S-CSM~\cite{51_S-BKI_gan2020}}&         {42.5} & {73.5} & {15.2} & {65.1} & {80.6} & {11.3} & {41.6} & {63.5} & {14.3}\\
        \textit{S-BKI~\cite{51_S-BKI_gan2020}}&         {41.9} & {78.5} & {18.1} & {65.6} & {85.8} & {18.0} & {40.2} & {75.3} & {20.3}\\
        \textit{SEE-CSOM~\cite{88_SEE-CSOM_deng2023}}&       {41.3} & {79.0} & {14.5} & {64.8} & {85.8} & {10.8} & {39.4} & {76.5} & {14.3}\\
        \textit{ConvBKI~\cite{47_ConvBKI2_wilson2023}}&  {42.3} & {80.2} & {14.2} & {68.4} & {86.7} & {9.5} & {41.9} & {71.1} & {13.2}\\
        \textit{\EBS~\cite{EBS_kim2024}} &                 {44.0} & {80.4} & {12.9} & {66.9} & {86.4} & {9.3} & {42.0} & {78.1} & {14.7}\\
        
        \midrule
        \textit{Ours (BEV)} &                                   \secnd{46.5} & \secnd{82.9} & \secnd{12.7} & \secnd{70.2} & \secnd{89.5} & \secnd{8.0} & \secnd{43.5} & \secnd{82.5} & \secnd{10.5} \\
        \textit{Ours (Cont.)} &                                 \first{46.7} & \first{83.1} & \first{12.6} & \first{70.7} & \first{89.6} & \first{7.9} & \first{43.6} & \first{82.6} & \first{10.4}\\
        
        \bottomrule
    \end{tabular}
    }
}
\vspace{-0.05in}
\end{table}

\begin{table}[t!]
\centering
\scriptsize
\renewcommand{\arraystretch}{1.0}
\caption{Ablation study results. Each row indicates the mapping performance after cumulative removal of components.}
\vspace{-0.05in}
\label{tab:ablation}
\normalsize{
\resizebox{1.0\linewidth}{!}{%
        \begin{tabular}{l c ccc}
        \toprule
        \multicolumn{1}{c}{\textbf{Component(s)}} & \textbf{Section} & \mIoU & \Acc & \BS$\downarrow$ \\
        \midrule
        \multicolumn{1}{l}{Ours} & -- & \first{47.3} & \first{83.5} & \first{13.0} \\ 
        {\small \textit{- Gaussian Refinement (Pruning)}} & \ref{Method:Refine} & 45.9 & 83.2 & 13.9 \\
        {\small \textit{- Gaussian Refinement (Merging)}} & \ref{Method:Refine} & 45.8 & 83.2 & 14.3 \\
        {\small \textit{- Anisotropic Gaussian Primitive}} & \ref{Method:Init} & 45.5 & 83.1 & 14.4 \\
        {\small \textit{- Isotropic Gaussian Primitive} (\EBS~\cite{EBS_kim2024})}  &  \ref{Method:Init} & 45.1 & 80.9 & 13.2 \\ 
        {\small \textit{- Evidential BKI} (S-BKI~\cite{51_S-BKI_gan2020})}  &  \ref{Method:EDL} & 43.8 & 78.9 & 16.3 \\ 
        \bottomrule
    \end{tabular}
    }
}
\vspace{-0.1in}
\end{table}

\subsection{Ablation Studies}
We conduct ablation studies on \RELLIS to isolate the contribution of each component in our framework, as summarized in \Tref{tab:ablation}. Starting from the S-BKI baseline, \EBS addresses semantic uncertainty and demonstrates substantial improvements. The introduction of isotropic Gaussian primitives transitions the framework to primitive-based operations, allowing local observations to be aggregated into compact, semantically coherent units. Extending primitives to anisotropic Gaussians enables geometry-aligned kernels, mitigating spatial uncertainty. Local merging improves mapping stability and computational efficiency, while adaptive pruning significantly enhances semantic performance.

\subsection{Runtime Analysis}
Finally, we evaluate the computational efficiency of our framework, averaging over five runs on a laptop with an Intel i7-12700H CPU. Gaussian construction achieves $9.73 \pm 0.21$ Hz, and evidential ellipsoidal BKI operates at $12.10 \pm 0.23$ Hz. The full framework runs at $5.39 \pm 0.11$ Hz, faster than the S-BKI baseline ($3.15 \pm 0.04$ Hz). This advantage arises from (1) the abstraction of individual points into Gaussians, which reduces the number of BKI computations, and (2) local merging, which further reduces redundant computation, as demonstrated by the throughput drop to $2.70 \pm 0.08$ Hz without refinement steps.

\section{CONCLUSIONS}
We propose an uncertainty-aware semantic mapping framework that addresses the limitations of S-BKI through anisotropic Gaussian primitives. Our approach tackles semantic uncertainty through evidential deep learning, spatial uncertainty through ellipsoidal kernels, and observation uncertainty through local context aggregation and pruning. By representing local geometry and semantics through Gaussian primitives, our method enables uncertainty-aware semantic propagation of sparse and noisy observations. Extensive experiments demonstrate superior performance in mapping quality and uncertainty calibration. The framework's robustness and versatility across different map representations highlight its practicality for autonomous systems. Future work includes improving uncertainty estimation to obtain more calibrated predictive uncertainties, refining the formal definitions of the uncertainties, and extending the uncertainty-aware mapping framework to open-set settings.

\addtolength{\textheight}{0cm}
                                  
\bibliographystyle{IEEEtran}
\bibliography{mybib}

\input{Appendix}
\end{document}

%% file: commands.tex
\newcommand{\balpha}{\boldsymbol{\alpha}}
\newcommand{\bmu}{\boldsymbol{\mu}}
\newcommand{\bSigma}{\boldsymbol{\Sigma}}
\newcommand{\hbthm}{\hat{\boldsymbol{\theta}}_m}
\newcommand{\hth}{\hat{\theta}}
\newcommand{\bpn}{\mathbf{p}_n}
\newcommand{\byn}{\mathbf{y}_n}
\newcommand{\bxn}{\mathbf{x}_n}
\newcommand{\hbxm}{\mathbf{\hat{x}}_m}
\newcommand{\argmax}{\mathrm{argmax}}
\newcommand{\indicator}{\mathbb{1}}
\definecolor{puddle}{rgb}{0.0, 0.0, 1.0}
\definecolor{object}{rgb}{0.8, 0.6, 1.0}
\definecolor{paved}{rgb}{1.0, 1.0, 0.0}
\definecolor{unpaved}{rgb}{1.0, 0.6, 0.8}
\definecolor{dirt}{rgb}{0.6, 0.298039, 0.0}
\definecolor{grass}{rgb}{0.435294, 1.0, 0.290196}
\definecolor{vegetation}{rgb}{0.0, 0.4, 0.0}

\definecolor{road}{rgb}{0.502, 0.251, 0.502}
\definecolor{sidewalk}{rgb}{0.957, 0.137, 0.910}
\definecolor{building}{rgb}{0.275, 0.275, 0.275}
\definecolor{fence}{rgb}{0.745, 0.600, 0.600}
\definecolor{pole}{rgb}{0.600, 0.600, 0.600}
\definecolor{sign}{rgb}{0.863, 0.863, 0.000}
\definecolor{kittivegetation}{rgb}{0.420, 0.557, 0.137}
\definecolor{terrain}{rgb}{0.596, 0.984, 0.596}
\definecolor{person}{rgb}{0.863, 0.078, 0.235}
\definecolor{car}{rgb}{0.000, 0.000, 0.557}
\definecolor{truck}{rgb}{0.000, 0.000, 0.275}
\definecolor{motorcycle}{rgb}{0.000, 0.000, 0.902}

\newcommand{\first}[1]{\textbf{#1}}
\newcommand{\secnd}[1]{\underline{#1}}
\newcommand\semcolor[1][black]{\fcolorbox{black}{#1}{\rule{0mm}{1mm}\rule{1mm}{0mm}}}

\newcommand{\calL}{\mathcal{L}}

\newcommand{\Dkl}{D_{\text{KL}}}
\newcommand{\dataset}{\mathcal{D} = \{(\bxn, \byn)\}^N_{n=1}}
\newcommand{\Uthres}{\mathrm{U}_{\text{thr}}}
\newcommand{\softmax}{\text{SoftMax}(\cdot)}

\newcommand{\dol}{\frac{d}{\ell}}
\newcommand{\tpk}{\text{TP}_c}
\newcommand{\fpk}{\text{FP}_c}
\newcommand{\fnk}{\text{FN}_c}

\newcommand{\Tref}[1]{Table~\ref{#1}}
\newcommand{\Fref}[1]{Fig.~\ref{#1}}
\newcommand{\Sref}[1]{Section~\ref{#1}}

\newcommand{\EBS}{EBS\xspace}
\newcommand{\RELLIS}{\textit{RELLIS-3D}\xspace}
\newcommand{\OFFROAD}{\textit{OffRoad}\xspace}
\newcommand{\KITTI}{\textit{KITTI-360}\xspace}
\newcommand{\IoU}{$\mathrm{IoU}$\xspace}
\newcommand{\mIoU}{$\mathrm{mIoU}$\xspace}
\newcommand{\Acc}{$\mathrm{Acc}$\xspace}
\newcommand{\BS}{$\mathrm{BS}$\xspace}
\newcommand{\ECE}{$\mathrm{ECE}$\xspace}

%% file: Appendix.tex
\newcommand{\Dir}{\text{Dir}}
\vspace{-0.2in}
\section*{Appendix}
\vspace{-0.05in}
\noindent To provide a more comprehensive view of our framework, we include extended details and additional results in this appendix. It covers supplementary material including background on related methods, explanations of our approach, and experimental analyses.

\section{Background: Evidential Semantic Mapping with Bayesian Kernel Inference}
\subsection{Semantic Bayesian Kernel Inference}
Semantic Bayesian Kernel Inference~(S-BKI)~\cite{51_S-BKI_gan2020} serves as the baseline for continuous semantic mapping frameworks. Given semantic points $\dataset$ with 3D coordinates $\bxn \in \mathbb R^3$ and one-hot labels $\byn \in \{0, 1\}^C$ over $C$ categories, S-BKI establishes a theoretical framework for continuous semantic inference at arbitrary query points. In practice, this framework is instantiated as voxel-based semantic mapping, where the environment is discretized into voxels at a predefined resolution and the center of each voxel is treated as a query point.

The semantic voxels are represented as $\{ (\hbxm, \hbthm) \}_{m=1}^M$, with each query point $\hbxm$ being the center of voxel $m$. At each voxel center, the semantic state is modeled by a Categorical distribution $\hbthm = [\hth_m^1, \ldots, \hth_m^C]$ satisfying $\sum^C_{c=1} \hth_m^c = 1$ and $\hth_m^c \in [0,1]$. Unlike conventional methods that rely only on measurements within each voxel, S-BKI incorporates neighboring points to estimate $\hbthm$, enabling more complete and spatially consistent predictions. Formally, the posterior distribution $p(\hbthm | \hbxm, \mathcal{D})$ is modeled within a probabilistic framework:
\begin{align}
\label{aBKI_derivation}
    \underbrace {\strut p(\hbthm | \hbxm, \mathcal{D}) }_{\text{posterior}}
        &\propto p(\mathcal{D} | \hbthm, \hbxm) p(\hbthm | \hbxm) \nonumber \\
        &\propto 
            \Big[ \prod^N_{n=1} 
                \underbrace {\strut p(\byn | \bxn, \hbthm, \hbxm) }_{\text{extended likelihood}} 
            \Big]
            \underbrace{\strut p(\hbthm | \hbxm)}_{\text{prior}} \nonumber\\
        &\propto
            \Big[ \prod^N_{n=1} 
                p(\byn | \hbthm)^{k(\hbxm, \bxn)}
            \Big]
            p(\hbthm | \hbxm) \nonumber \\
        &\propto
            \Big[ \prod^N_{n=1} 
                \bigl(
                    \prod^C_{c=1} (\hth_m^c)^{y_n^c} 
                \bigr)^{k(\hbxm, \bxn)}
            \Big] \prod^C_{c=1} (\hth_m^c)^{\alpha_0^c - 1} \nonumber \\
        &\propto
            \prod^C_{c=1} (\hth_m^c)^{\alpha_0^c + \sum^N_{n=1} k(\hbxm, \bxn)y_n^c - 1},
\end{align}
where $k(\cdot, \cdot)$ is a kernel function that satisfies the following conditions:
\begin{equation}
\begin{aligned}\label{aBKI_kernel_requirement}
    k(\mathbf x, \mathbf x) = 1\ \ ^{\forall} \mathbf x \quad\text{and}\quad k(\mathbf x, \mathbf x') \in [0, 1]\ \ ^{\forall} \mathbf x, \mathbf x',
\end{aligned}
\end{equation}
and the Dirichlet prior $\Dir(C, \boldsymbol{\alpha}_0)$ is adopted to enable incremental Bayesian inference since it is the conjugate prior of the Categorical distribution. Accordingly, the posterior $\Dir(C, \boldsymbol{\alpha}_{m})$ for query $\hbxm$ can be recursively updated as:
\begin{equation}
\begin{aligned}\label{aBKI_final_update}
    \alpha^c_{m, t} \leftarrow \alpha_{m, t-1}^c + \sum^N_{n=1} k(\hbxm, \bxn) \cdot y^c_n,\quad \alpha^c_{m, 0} = \alpha^c_0,
\end{aligned}
\end{equation}
where $\alpha_0^c \in \mathbb R^+$ is an initial hyperparameter for each class and $\boldsymbol{\alpha}_{m,t}$ denotes posterior parameters at time $t$. The expectation and variance of $\hbthm$ given the posterior Dirichlet parameters $\boldsymbol{\alpha}_m$ are calculated as:
\begin{equation}
\begin{aligned}\label{aBKI_final_variance}
    S_{m} = \sum^C_{c=1} \alpha_{m}^c, \ 
    \mathbb{E}[\hth_{m}^c] = \frac{\alpha_{m}^c}{S_{m}}, \ 
    \mathrm{Var}[\hth_{m}^c] = \frac{\alpha_{m}^c (S_{m} - \alpha_{m}^c)}{S_{m}^2(S_{m} + 1)} .
\end{aligned}
\end{equation}
The predicted semantic label $\psi_m$ for voxel $m$ is the class with the largest posterior mean,
\begin{equation}
\begin{aligned}\label{aBKI_voxel_semantic}
    \psi_m = \arg\max_c \mathbb{E}[\hth_m^c] = \arg\max_c \frac{\alpha_{m}^c}{S_{m}} = \arg\max_c \alpha_{m}^c .
\end{aligned}
\end{equation}
The variance $\mathrm{Var}[\hth_m^{\psi_m}]$ serves as a proxy for the uncertainty of the estimate\footnote{For brevity and readability, $\psi_m$ is denoted as $\psi$ in the main text.}~\cite{51_S-BKI_gan2020, 47_ConvBKI2_wilson2023}.

\begin{figure}[t]
\centering
\includegraphics[width=0.75\linewidth]{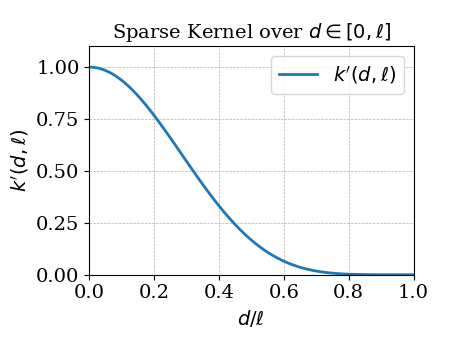}
\vspace{-0.175in}
\caption{The sparse kernel $k'(d,\ell, \sigma_0=1)$ decays as the ratio $d/\ell$ increases.}
\label{fig:appendix_sparse_kernel}
\vspace{-0.175in}
\end{figure}

As each observation is weighted by the kernel $k(\cdot, \cdot)$, its design determines both the spatial influence and efficiency of updates. A common choice is to let $k(\cdot, \cdot)$ decay with the distance $d = \|\hbxm - \bxn\|$ and truncate beyond a length scale $\ell$, thereby emphasizing nearby points while suppressing distant ones. In practice, the sparse kernel~\cite{66_SparseKernel_melkumyan2009} is widely adopted in continuous mapping~\cite{49_BGKOctoMap_doherty2017, 51_S-BKI_gan2020, 88_SEE-CSOM_deng2023} (\Fref{fig:appendix_sparse_kernel}):
\begin{equation}\label{aBKI_5_reparam}
\begin{aligned}
    k(\hbxm, \bxn) &= k'(d, \ell, \sigma_0) \\
                &=\underset{d < \ell}{\indicator} \: \sigma_0 \Big[ \frac{2 + \cos (2\pi \dol)}{3} (1 - \dol) + \frac{1}{2\pi} \sin (2\pi \dol) \Big] ,
\end{aligned}
\end{equation}
where $\indicator$ is the indicator function, and $\sigma_0$ is set to $1$ to satisfy \eqref{aBKI_kernel_requirement}. Since the sparse kernel decays smoothly to zero at $d = \ell$, the length scale $\ell$ confines the spatial support of BKI updates.

\subsection{Evidential Deep Learning}
To incorporate predictive uncertainty into the mapping process, we adopt the Evidential Deep Learning~(EDL) framework~\cite{10_EDL_sensoy2018} for semantic segmentation. Given a 2D dataset $\mathcal D_\text{2D} = \{(\mathbf X_i, \mathbf Y_i)\}_{i=1}^{N_{\text{2D}}}$ with images $\mathbf X_i \in \mathbb R^{H \times W \times 3}$ and pixel-wise labels $\mathbf Y_i \in \{0,1\}^{H\times W\times C}$, EDL enables direct estimation of both class probabilities and their uncertainty. Importantly, the same formulation can also be applied to 3D semantic segmentation, where a dataset $\mathcal D_\text{3D}=\{(\mathbf X_l,\mathbf Y_l)\}_{l=1}^{N_{\text{3D}}}$ consists of point clouds $\mathbf X_l \in \mathbb R^{L\times 4}$ with per-point semantic labels $\mathbf Y_l \in \{ 0,1 \}^{L\times C}$. In this work, we perform 2D semantic segmentation to leverage rich semantic information while considering annotation efficiency. For notational convenience, we denote an arbitrary pixel and its corresponding label in $(\mathbf X_i, \mathbf Y_i)$ by $(\mathbf x_i, \mathbf y_i)$.

Unlike conventional classifiers that rely on a $\softmax$ layer to output class probabilities, EDL predicts non-negative evidence $\mathbf{\tilde{e}}_i = [\tilde{e}_i^1, \ldots, \tilde{e}_i^C]$ for each $\mathbf x_i$, parameterizing a Dirichlet $\Dir(C, \tilde{\balpha}_i)$ with $\tilde{\alpha}_i^c=\tilde{e}_i^c+1$. EDL yields both class probabilities and a predictive uncertainty via the belief mass $\mathbb{\tilde{M}}_i = (\{ \tilde{b}^c_i \}^C_{c=1}, \tilde{u}_i)$:
\begin{equation}\label{aEDL_belmass}
\begin{aligned}
    \tilde{b}_i^c = \frac{\tilde{e}_i^c}{\tilde{S}_i},\quad\tilde{u}_i = \frac{C}{\tilde{S}_i},\quad\text{where } \tilde{u}_i + \sum_{c=1}^C \tilde{b}_i^c = 1 .
\end{aligned}
\end{equation}
Here, $\tilde{b}_i^c$ is the belief for the class $c$, $\tilde{S}_i = \sum^C_{c=1}\tilde{\alpha}_i^c$ denotes the total evidence strength aggregated across classes, and $\tilde{u}_i$ quantifies the overall predictive uncertainty. The expected class probability is
\begin{equation}\label{aEDL_belprob}
\begin{aligned}
    \tilde{p}_i^c = \tilde{b}_i^c + \frac{\tilde{u}_i}{C} ,
\end{aligned}
\end{equation}
which shows that the class probability can be decomposed into a belief component and an evenly distributed share of the uncertainty mass.

While the formulation above describes how evidence is transformed into class probabilities and uncertainty, the network must also be trained to produce meaningful evidence. To this end, we adopt the loss function from~\cite{10_EDL_sensoy2018, 13_EvPSNet_sirohi2023} to guide the network to increase the evidence for the ground truth class while suppressing misleading evidence in other classes. In practice, this is achieved by combining the evidential loss $\calL_{\text{EDL}}$, which encourages larger evidence for the correct class relative to the total evidence, with a KL regularization term $\calL_{\text{reg}}$, which penalizes misleading evidence:
\begin{equation}\label{aEDL_evidential_loss}
\begin{aligned}
    \calL_{\text{EDL}} (\mathbf{x}_i, \mathbf y_i) &= \sum^C_{c=1} y_i^c (\log(\tilde{S}_i) - \log(\tilde{\alpha}_i^c)), \\
    \calL_{\text{reg}} (\mathbf{x}_i, \mathbf y_i) &= \Dkl\Big[\Dir\left(\mathbf{\tilde{p}}_i | \boldsymbol{\hat{\alpha}}_i \right)\,\|\,\Dir\left(\mathbf{\tilde{p}}_i | [ 1, \ldots, 1 ] \right) \Big], 
\end{aligned}
\end{equation}
where the modified Dirichlet parameters $\boldsymbol{\hat{\alpha}}_i$ are obtained by setting the ground truth class parameter to $1$ and keeping all other parameters unchanged, so the KL term penalizes only evidence assigned to incorrect classes:
\begin{equation}
\begin{aligned}
    \hat{\alpha}_i^c = 
        &\begin{cases}
            1 &\text{if } y_i^c = 1 \\
            \tilde{\alpha}_i^c &\text{if } y_i^c = 0 \\
        \end{cases} .
\end{aligned}
\end{equation}
The overall training objective is then defined as:
\begin{align}\label{aEDL_OVERALL_LOSS}
    \calL(\mathbf{x}_i, \mathbf y_i) = \calL_{\text{EDL}}(\mathbf{x}_i, \mathbf y_i) + \lambda_{\text{KL}} \calL_{\text{reg}}(\mathbf{x}_i, \mathbf y_i) , 
\end{align}
where $\lambda_{\text{KL}}$ is a KL annealing coefficient.

Once the EDL model is trained to minimize the loss in \eqref{aEDL_OVERALL_LOSS}, the probability $\mathbf{\tilde{p}}_i$ and its uncertainty $\tilde{u}_i$ are reliably estimated for each input $\mathbf{x}_i$. Incorporating these estimates into the mapping process can enhance the performance of semantic mapping, even under perceptually challenging off-road environments.

\subsection{Evidential Semantic Mapping}
Although S-BKI provides an effective framework for continuous semantic mapping, it assumes equal reliability across all input points, neglecting the varying uncertainty inherent in semantic predictions. To address this limitation, we introduced EBS~\cite{EBS_kim2024}, an evidential semantic mapping framework that extends S-BKI by explicitly incorporating predictive uncertainty via Evidential Deep Learning. Specifically, EBS adopts probabilistic semantic representations and an adaptive kernel, resulting in more robust and consistent semantic fusion.

To incorporate uncertainty into the mapping process, EBS generalizes the input semantic points $\dataset$ into evidential points $\mathcal{D'} = \{ (\bxn, \bpn, u_n) \}^N_{n=1}$, where each point consists of a 3D coordinate $\bxn$, a probability distribution $\bpn$ over the $C$ semantic classes, and a corresponding uncertainty estimate $u_n$. Leveraging the generalized probabilistic input $\bpn$, the likelihood at query point $\hbxm$ can be formulated naturally using the Continuous Categorical distribution~\cite{89_gordon2020}:
\begin{align}
  p(\bpn | \hbthm) \propto \prod^C_{c=1} (\hth_m^c)^{p_n^c} .
\end{align}
Because the Dirichlet distribution is conjugate to both the Categorical and Continuous Categorical likelihoods, the posterior retains the same analytical form as~\eqref{aBKI_derivation}:
\begin{align}
\label{auBKI_derivation}
    p(\hbthm | \hbxm, \mathcal{D'}) 
        &\propto
            \prod^C_{c=1} (\hth_m^c)^{\alpha_0^c + \sum^N_{n=1} k(\hbxm, \bxn)p_n^c - 1}.
\end{align}
Consequently, the update rule in \eqref{aBKI_final_update} is generalized to incorporate the full distribution $\bpn$, replacing the discrete one-hot label $\byn$. This enables a soft fusion of semantic evidence, where each point contributes to multiple classes based on its predictive confidence, thereby better capturing the uncertainty in perceptual predictions.
\begin{equation}
\begin{aligned}\label{auBKI_final_update}
  \alpha^c_{m,t} \leftarrow \alpha_{m,t-1}^c + \sum^N_{n=1} k(\hbxm, \bxn) \cdot p^c_n,\quad \alpha^c_{m, 0} = \alpha^c_0 .
\end{aligned}
\end{equation}

To further leverage uncertainty estimates, EBS introduces an adaptive kernel that modulates the spatial influence of each observation according to its reliability. Building upon the sparse kernel in \eqref{aBKI_5_reparam}, the adaptive formulation takes the form:
\begin{equation}
  \begin{aligned}\label{auBKI_kernel}
      k(\hbxm, \bxn) =
          &\begin{cases}
              k'(d, \ell \cdot \beta e^{1 - \gamma u_n}, \sigma_0) &\text{if } u_{n} \le \Uthres \\
              0 &\text{if } u_n > \Uthres ,
          \end{cases}
  \end{aligned}
\end{equation}
where $\beta$ and $\gamma$ are scaling hyperparameters\footnote{To simplify the parameterization, we fix $\gamma = 1$, and we set $\sigma_0 = 1$ to satisfy \eqref{aBKI_kernel_requirement}.}. This adaptive kernel discards highly uncertain observations exceeding a dynamic threshold $\Uthres$, which filters out the top $\tilde u$ percentile of each input. The spatial influence of the remaining observations is then scaled according to their uncertainty, enabling more confident predictions to contribute over a broader region. This strategy of filtering outliers and adaptively weighting contributions yields a more reliable semantic map.

\section{Evidential Ellipsoidal BKI}
\subsection{Gaussian Initialization (\Sref{Method:Init})}
\noindent\textbf{Clustering}
Our framework performs clustering to construct Gaussian primitives from the evidential points $\mathcal{D'} = \{ (\bxn, \bpn, u_n) \}_{n=1}^{N}$. For clustering stability, the points are first partitioned by their most likely semantic label:
\begin{equation}\label{aGaussian_partition}
\mathcal{D}'_c = \{ (\bxn, \bpn, u_n) \in \mathcal{D'} \;|\; \arg\max_{c'} p_n^{c'} = c \}.
\end{equation}
Clustering is thus performed independently within each semantic class, as we empirically found this strategy to yield more stable and reliable performance. Within each partition $\mathcal{D}'_c$, we apply K-Means++ clustering~\cite{267_KMenasPP_arthur2006} to form spatially coherent subsets $\{ \mathcal{C}_{j} \}_{j=1}^{J_c}$ where $J_c$ denotes the number of clusters for class $c$. We then pool the clusters across all classes into a unified set $\mathbb{C}=\{ \mathcal{C}_{j} \}_{j=1}^J$ with $J = \sum_{c=1}^C J_c$, removing the class index for subsequent Gaussian construction.

\noindent\textbf{Aggregation} 
Each cluster $\mathcal{C}_{j}$ is then converted into a Gaussian primitive $\mathcal G_j = (\bmu_j, \bSigma_j, \mathbf p_j, u_j)$, where the geometric component $(\bmu_j, \bSigma_j)$ models the local geometry through mean $\boldsymbol{\mu}_j$ and covariance $\bSigma_j$, and the semantic component $(\mathbf p_j, u_j)$ represents the evidential beliefs. 

For efficient merging, we employ the Method of Moments (MoM)~\cite{268_MoM_scott2001} to represent each Gaussian. MoM represents each Gaussian using the unnormalized first moment $\mathbf m^{(1)}_j$, the unnormalized second moment $\mathbf M^{(2)}_j$, and a normalization constant $w_j$. These quantities can be converted back to Gaussian parameters as:
\begin{equation}\label{aGaussian_mom2gauss}
 \boldsymbol{\mu}_j = \frac{1}{w_j}\mathbf{m}^{(1)}_j,\quad \bSigma_j = \frac{1}{w_j}\mathbf{M}^{(2)}_j - \boldsymbol{\mu}_j\boldsymbol{\mu}_j^\top .
\end{equation}
Using this representation, Gaussian primitives can be efficiently merged by directly accumulating their moments. Specifically, when two primitives $\mathcal G_A$ and $\mathcal G_B$ are merged to yield $\mathcal G$, the merged normalization constant, the first moment, and the second moment are obtained as:
\begin{equation}\label{aGaussian_momMerge}
\begin{gathered}
\mathbf m^{(1)} = \mathbf m^{(1)}_A + \mathbf m^{(1)}_B, \quad
\mathbf M^{(2)} = \mathbf M^{(2)}_A + \mathbf M^{(2)}_B, \\
w = w_A + w_B .
\end{gathered}
\end{equation}
The resulting Gaussian parameters $\boldsymbol{\mu}$ and $\bSigma$ are then recovered using \eqref{aGaussian_mom2gauss}. This moment-based merging is computationally efficient and well-suited for incremental map construction.

Meanwhile, the semantic component $(\mathbf p_j, u_j)$ is obtained by aggregating the semantic predictions $(\mathbf p_n, u_n)$ from all points in $\mathcal C_j$ using the DST-based combination rules~\cite{72_TMCJournal_han2022}. In this framework, each prediction is expressed as a belief mass 
$\mathbb M = \bigl(\{ b^c \}_{c=1}^C,\, u \bigr)$, where $b^c$ is the belief mass for class $c$ and $u$ represents the overall uncertainty. Given two belief masses $\mathbb M_A = \bigl(\{ b^c_A \}_{c=1}^C,\, u_A \bigr)$ and $\mathbb M_B = \bigl(\{ b^c_B \}_{c=1}^C,\, u_B \bigr)$, their combination $\mathbb M = \mathbb M_A \oplus \mathbb M_B$ is defined as:
\begin{equation}\label{aGaussian_dstcomb}
    b^c = \frac{1}{1-\eta} (b^c_A b^c_B + b^c_A u_B + b^c_B u_A), \quad u = \frac{1}{1-\eta}u_A u_B,
\end{equation}
where $\eta = \sum_{x \ne y} b^x_A b^y_B$ quantifies the conflict between two belief masses. To apply the rule in \eqref{aGaussian_dstcomb}, the evidential points are converted into belief masses using \eqref{aEDL_belmass} and \eqref{aEDL_belprob}.

\subsection{Gaussian Refinement (\Sref{Method:Refine})}
Although Gaussian initialization provides structured local representations, treating each frame independently introduces several drawbacks: (i) the lack of temporal consistency prevents effective use of information across frames, (ii) redundant primitives accumulate over time, increasing computational overhead, and (iii) sparse regions yield unstable estimates due to limited observations.

\noindent\textbf{Merging}
To improve geometric stability and computational efficiency, we merge locally consistent Gaussian primitives. For each Gaussian, all neighboring primitives within a radius $d_L$ are examined for semantic consistency. If they all belong to the same semantic class, those within the smaller radius $d_S$ are merged. By enforcing semantic homogeneity over the larger neighborhood $d_L$ prior to merging within $d_S$, we ensure that refinement occurs only in reliably uniform regions. As summarized in \Tref{table:hyperparams}, we set $d_L = 5\ell$ and $d_S = \ell$, which yields a conservative merging strategy and prevents excessive fusion.

When these conditions are satisfied, the geometric components $(\bmu_j, \bSigma_j)$ are updated using the MoM representation introduced in \eqref{aGaussian_mom2gauss} and \eqref{aGaussian_momMerge}. This formulation enables efficient in-place merging through moment accumulation~\cite{4_GMMap_li2024}. The semantic components are fused using the DST-based combination rule in \eqref{aGaussian_dstcomb}, ensuring robust integration of class probabilities and uncertainties.

\noindent\textbf{Pruning}  
To address semantic inconsistencies that mainly arise from uncertain distant observations (as shown in \Fref{fig:uncmodel}), we adopt a relative pruning strategy. Inspired by the human visual system, which maintains initial impressions until contradicted by more reliable evidence, we evaluate the reliability of each Gaussian primitive $\mathcal G_j$ and prune it when a conflicting but more reliable primitive $\mathcal G_i$ exists, as defined in \eqref{prune_cond}. The sensitivity of pruning is controlled by $\epsilon$, which can be tuned according to sensor or environmental characteristics; however, we fix $\epsilon = 2.5$ across all experiments and find this setting to be robust across diverse conditions.

\subsection{Gaussian as an Ellipsoid (\Sref{Method:Kernel})}
When BKI inference is formulated over Gaussian primitives rather than raw data points, a new notion of distance between a primitive and a query location becomes necessary. In conventional BKI, kernel inference operates on individual observations using Euclidean distance, whereas our framework introduces Gaussian primitives as an intermediate scene representation. Consequently, kernel inference must be defined between a Gaussian primitive and a query point.

An appropriate distance metric must capture not only the location $\bmu_j$ of a Gaussian primitive $\mathcal G_j$ but also its spatial extent and anisotropic geometry $\bSigma_j$. A naive choice, such as the Euclidean distance between the primitive mean $\boldsymbol{\mu}_j$ and the query point, fails to capture the anisotropic shape of the Gaussian. Probabilistic measures such as the Mahalanobis distance incorporate the covariance information but are less intuitive and require nontrivial parameter tuning. Therefore, we explicitly interpret each Gaussian primitive $\mathcal G_j$ as an anisotropic ellipsoid and define the distance as the shortest Euclidean projection from the query point to the ellipsoid boundary. Formally, the ellipsoid $\mathcal E_j$ is defined by a Mahalanobis distance threshold $\tau$ corresponding to a desired confidence level:
\begin{equation}\label{aGauss_Ellipsoid}
    \mathcal E_{j} = \{ \mathbf{x} \mid (\mathbf{x} - \boldsymbol{\mu}_{j})^\top \bSigma_{j}^{-1} (\mathbf{x} - \boldsymbol{\mu}_{j}) \le \tau \}.
\end{equation}

We choose $\tau$ to enclose a specified fraction $\tilde{\tau}$ (we use $10\%$) of the Gaussian’s probability mass. For a $d$-dimensional Gaussian, the squared Mahalanobis distance follows a chi-squared distribution with $d$ degrees of freedom, so $\tau$ is the chi-squared quantile at level $\tilde{\tau}$:
\begin{equation}\label{aGauss_tau}
\tau = F^{-1}_{\chi^2_d}(\tilde{\tau}),
\end{equation}
where $d=3$ for 3D voxel mapping and $d=2$ for 2D BEV mapping. Intuitively, $\tilde{\tau}$ specifies the fraction of the Gaussian probability mass enclosed by the ellipsoid (e.g., $\tilde{\tau}=0.10$ encloses $10\%$ of the mass).

The distance between a primitive $\mathcal G_j$ and a query point $\hbxm$, denoted by $d^\ast(\hbxm, \mathcal G_j) := d^\ast$, is defined as the shortest Euclidean distance from $\hbxm$ to the ellipsoidal surface of $\mathcal E_j$. The closest point $\mathbf v^\ast$ on the surface is obtained by solving the following constrained minimization:
\begin{equation}
d^\ast = \min_{\mathbf{v} \in \mathbb{R}^3} \|\hbxm - \mathbf{v}\|^2 \text{ s.t. } (\mathbf{v} - \boldsymbol{\mu}_{j})^\top \bSigma_{j}^{-1} (\mathbf{v} - \boldsymbol{\mu}_{j}) = \tau,
\end{equation}
with $d^\ast(\hbxm, \mathcal G_j) = 0$ when $\hbxm \in \mathcal E_j$.

\subsection{Uncertainty Decomposition}\label{Appendix:UncDec}
While our formulation in \eqref{E2BKI_final_update} enhances continuous semantic mapping, reliable uncertainty estimation $u_m$ for the posterior $\boldsymbol{\alpha}_m$ remains crucial. A common surrogate is the normalized Dirichlet variance $\mathrm{Var}[\hbthm^{\psi_m}]$, which monotonically decreases as evidence accumulates and can therefore underestimate uncertainty in semantically ambiguous regions. To mitigate overconfidence and provide reliable uncertainty estimates for safe decision-making, we decompose $u_m$ into semantic uncertainty $u_{m, \text{sem}}$ and evidence sparsity $u_{m, \text{spa}}$.

Concretely, the semantic uncertainty is computed as a kernel-weighted average of per-primitive uncertainties $u_j$:
\begin{equation}\label{E2BKI_unc_sem}
u_{m,\text{sem}} = \frac{\sum_{j=1}^{J^*} \tilde k(\hbxm,\mathcal G_j) \cdot u_j}{\sum_{j=1}^{J^*} \tilde k(\hbxm,\mathcal G_j)}.
\end{equation}
where $J^*$ is the total number of Gaussian primitives. To isolate evidence sparsity, we flatten the posterior Dirichlet distribution with $S_m = \sum_{c=1}^C \alpha_m^c$ to $[S_m/C,\,\ldots\,,S_m/C]$, where $C$ is the number of semantic classes. Plugging this into the variance formula in \eqref{aBKI_final_variance} yields:
\begin{align}\label{E2BKI_unc_spa}
u_{m,\text{spa}} &= \frac{\alpha_{m}^c (S_{m} - \alpha_{m}^c)}{S_{m}^2(S_{m} + 1)} \nonumber\\
                 &= \frac{C-1}{C^2(S_m + 1)} \nonumber\\
                 &= \frac{C-1}{C^2(\sum_{c=1}^C \alpha^c_0 + \sum_{j=1}^{J^*} \tilde k(\hbxm,\mathcal G_j) + 1)},
\end{align}
with $S_m = \sum_{c=1}^C \alpha^c_0 + \sum_{j=1}^{J^*} \tilde k(\hbxm,\mathcal G_j)$ from our update rule. Finally, the overall uncertainty $u_m$ is obtained by combining the semantic and spatial uncertainty components:
\begin{equation}\label{E2BKI_unc}
u_m  = {u}_{m,\text{sem}} + {u}_{m,\text{spa}}.
\end{equation}

\section{Experimental Details}
\subsection{Datasets}
\noindent\textbf{\RELLIS} provides RGB images ($1920{\times}1200$), Ouster OS1-64 LiDAR scans, and accurate robot poses with 2D semantic segmentation annotations. We preprocess the dataset to retain only samples with matched RGB–LiDAR–annotation triplets. After filtering, the five subsets contain $1{,}200$, $1{,}010$, $1{,}443$, $581$, and $860$ frames, respectively. We adopt a 5-fold evaluation protocol in which four subsets are used for training and the remaining one for testing in each fold. We repeat this process for all five folds and report the average across folds.

\noindent\textbf{\OFFROAD} consists of RGB images ($2464{\times}1713$), Ouster OS1-128 LiDAR scans, and estimated robot poses~\cite{123_METAVerse_seo2023} with manually annotated 2D semantic segmentation labels. LiDAR-camera projection utilizes an advanced extrinsic calibration procedure based on~\cite{CAL_kwak2011}, ensuring accurate 2D-3D correspondence for reliable semantic mapping evaluation. Collected through high-speed navigation under diverse conditions, including different seasons (spring, fall, and winter), lighting variations, and terrain types, this dataset presents significant challenges for reliable 3D mapping. The dataset comprises $8{,}893$ images for 2D semantic segmentation network training and $1{,}400$ frames ($300$, $400$, $300$, $400$ per sequence) for 3D semantic mapping evaluation.

\noindent\textbf{\KITTI} spans $73.7$\,km of urban driving sequences collected in Karlsruhe, Germany, using a Velodyne HDL-64E LiDAR and synchronized RGB cameras. The dataset provides high-precision GPS/IMU localization and 2D semantic annotations with 19 semantic classes. We use front-view camera images ($1408{\times}376$) and corresponding LiDAR point clouds. After filtering through RGB-LiDAR-annotation matching, we split the dataset into sequences 00 and 02-06 for training ($45{,}220$ frames); 07 and 09 for validation ($13{,}372$); and 10 for testing ($2{,}688$). This urban dataset poses distinct challenges, including dense traffic, multi-story buildings, and dynamic lighting.

\subsection{Implementation Details}
\noindent\textbf{2D Semantic Segmentation}
Semantic labels are remapped into 7 classes for off-road~(\Tref{tab:offroad_quant}) and 11 classes for urban settings~(\Tref{tab:urban_quant}). We use the lightweight LRASPP architecture~\cite{266_LRASPP_howard2019} in place of the previous DeepLabV3 architecture~\cite{56_DeepLabV3_chen2017, EBS_kim2024} to improve computational efficiency for real-time deployments while maintaining comparable semantic accuracy. The KL-annealing coefficient $\lambda_{\text{KL}}$ increases linearly during training as $\lambda_{\text{KL}} = \frac{1}{120}\cdot\text{epoch}$. We optimize with Adam $(\beta_1{=}0.5,\ \beta_2{=}0.999)$ and use a StepLR scheduler that multiplies the learning rate by $0.8$ every $5$ epochs. Additional training hyperparameters are summarized in \Tref{tab:a2dparams}. Note that our framework is agnostic to the uncertainty estimation method and works with any approach that produces semantic probabilities $\bpn$ and uncertainties $u_n$.

\begin{table}[t!]
\centering
\scriptsize
\renewcommand{\arraystretch}{0.75}
\caption{Hyperparameters for training semantic segmentation networks.}
\label{tab:a2dparams}
\normalsize{
\resizebox{0.8\linewidth}{!}{%
        \begin{tabular}{c ccc}
        \toprule
        & {\textbf{\RELLIS}} & {\textbf{\OFFROAD}} & {\textbf{\KITTI}} \\
        \midrule
        epoch & $15$ & $15$ & $10$ \\
        batch size & $16$ & $16$ & $24$ \\
        learning rate & $0.0002$ & $0.0002$ & $0.00004$ \\
        \bottomrule
    \end{tabular}
    }
}
\vspace{-0.2in}
\end{table}

\noindent\textbf{3D Semantic Mapping}
We adopt a single set of mapping hyperparameters across all datasets as summarized in \Tref{table:hyperparams}. The only dataset-specific setting is the number of K-Means++ clusters for initializing Gaussian primitives: 256 for \RELLIS and 1{,}024 for the other datasets, to balance map compactness and geometric detail.

\subsection{Evaluation Setup}
Dense ground-truth occupancy maps are rarely available in real-world datasets and are costly to construct. To reflect online operation under limited sensing, we therefore build maps using every fifth frame and evaluate them against reference point-cloud queries constructed from the full sequence. Accordingly, the $20\%$ data availability in \Fref{fig:robust:a} represents our primary evaluation setting. To construct the ground-truth queries, we accumulate all frames, voxelize the aggregated cloud at one quarter of the evaluation resolution ($0.05$\,m), and retain only voxels whose points exhibit consistent semantics.

\noindent\textbf{Evaluation Metrics} For evaluation, semantic quality is reported with class-wise Intersection over Union (\IoU) and its mean (\mIoU), and we additionally report an accuracy metric (\Acc) that couples geometric occupancy with semantic correctness. Let $\tpk, \fpk$, and $\fnk$, represent the number of true positive, false positive, and false negative queries for class $c$, respectively. Then, semantic quality metrics are computed as follows:
\begin{align}
    \text{\IoU}_c &= \frac{\tpk}{\tpk+\fpk+\fnk},\quad\text{\mIoU} = \frac{1}{C} \sum_{c=1}^C \text{\IoU}_c,
\end{align}
\begin{align}
    \text{\Acc} &= \frac{1}{Q} \sum_{c=1}^{C} \tpk,
\end{align}
where $Q$ is the total number of ground-truth queries. For uncertainty calibration we use the Brier Score (\BS$\downarrow$) and the Expected Calibration Error (\ECE$\downarrow$):
\begin{align}
    \mathrm{BS} &= \frac{1}{Q'} \sum^{Q'}_{m=1} (\indicator_{\text{correct}, m} - \mathrm{Conf}_m)^2, \nonumber\\
    \mathrm{ECE} &= \sum^{B}_{b=1} \frac{|\mathcal B_b|}{Q'} | \mathrm{Acc}(\mathcal B_b) - \mathrm{Conf}(\mathcal B_b) |,
\end{align}
where $Q'$ is the number of occupied queries, and $\indicator_{\text{correct}, m}$ is an indicator function that equals $1$ if the predicted semantic label of the $m$-th query matches the ground truth and $0$ otherwise. The confidence of the $m$-th query is defined as $\mathrm{Conf}_m = 1 - \widehat{\mathrm{Var}}[\hth_m^{\psi_m}]$ for variance-based baselines, where $\widehat{\mathrm{Var}}[\hth_m^{\psi_m}]$ denotes the normalized variance obtained from $\mathrm{Var}[\hth_m^{\psi_m}]$ in \eqref{aBKI_final_variance}. For our method, the confidence is given by $\mathrm{Conf}_m = 1 - u_m$, as defined in \eqref{E2BKI_unc}.

For \ECE, the confidence values are partitioned into $B$ disjoint bins $\{\mathcal B_b\}_{b=1}^B$ according to their magnitude. For each bin $\mathcal B_b$, $\mathrm{Acc}(\mathcal B_b)$ denotes the empirical accuracy
\footnote{$\mathrm{Acc}(\mathcal B_b)$ is the bin-wise accuracy and is distinct from the overall metric \Acc.}
and $\mathrm{Conf}(\mathcal B_b)$ the average confidence of queries $\mathrm{Conf}_m$ assigned to that bin $\mathcal B_b$. The ECE measures the expected discrepancy between accuracy and confidence across all bins, weighted by the proportion of queries in each bin. In all experiments, we use $B = 15$.

\begin{figure*}[t]
\begin{center}
\includegraphics[width=\linewidth]{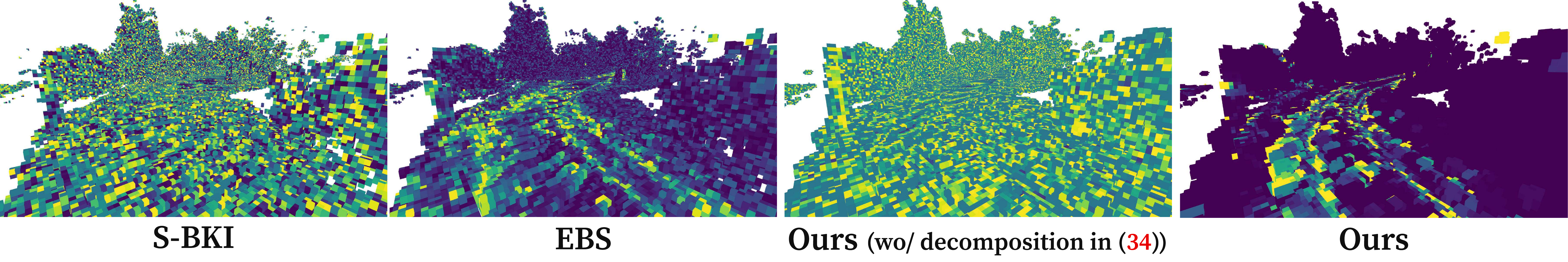}
\end{center}
\caption{Uncertainty map comparison for the fourth-row scene of \Fref{fig:mainqualitative}.}
\label{fig:avarmap}
\end{figure*}

\begin{figure}[t]
\centering
\includegraphics[width=\linewidth]{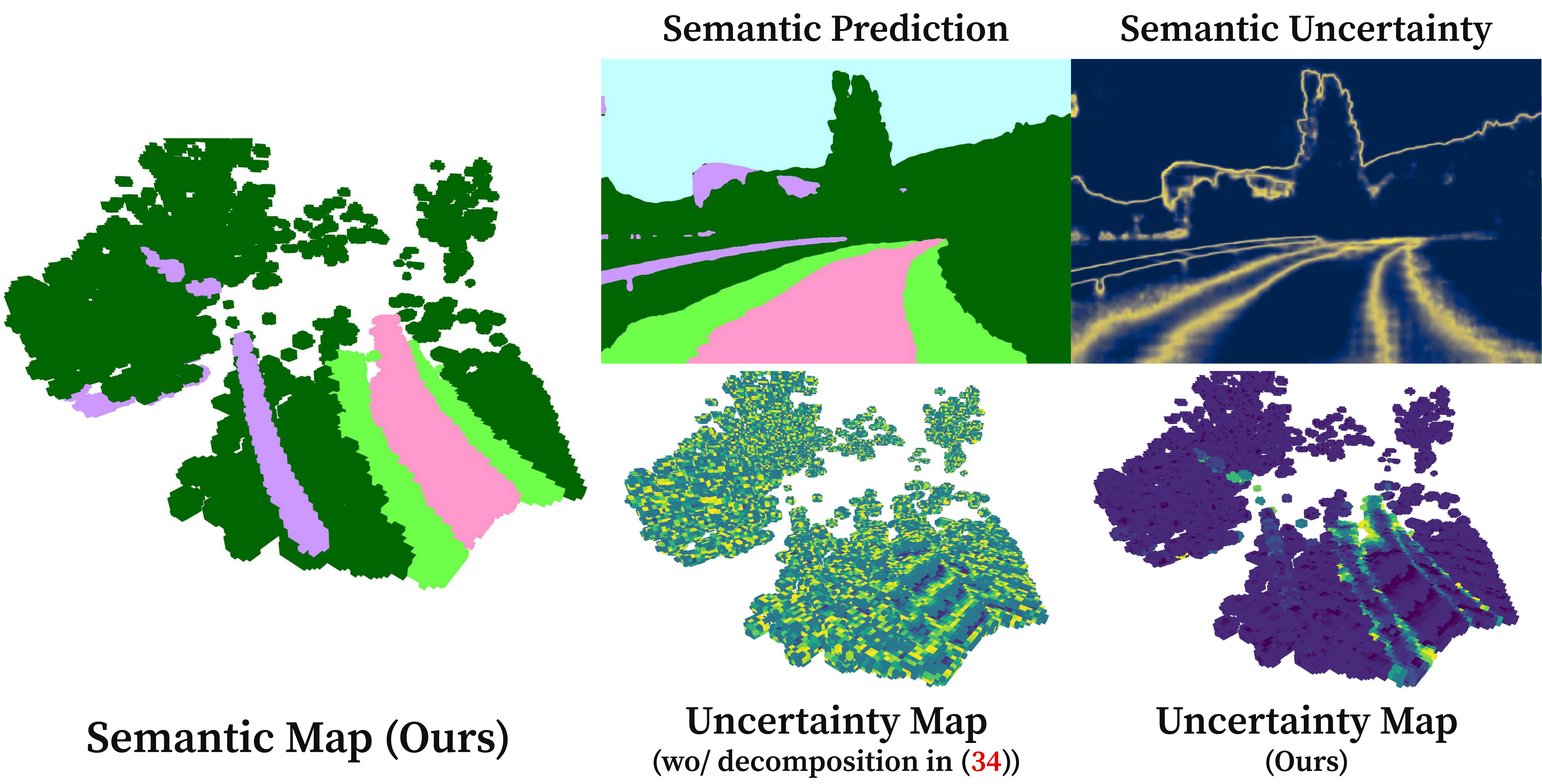}
\caption{Comparison of uncertainty maps on the same scene as \Fref{fig:avarmap}. Note that uncertainty decomposition does not affect the semantic map.}
\label{fig:avarmapone}
\vspace{-0.1in}
\end{figure}

\begin{figure*}[t]
\begin{center}
\includegraphics[width=1.0\textwidth]{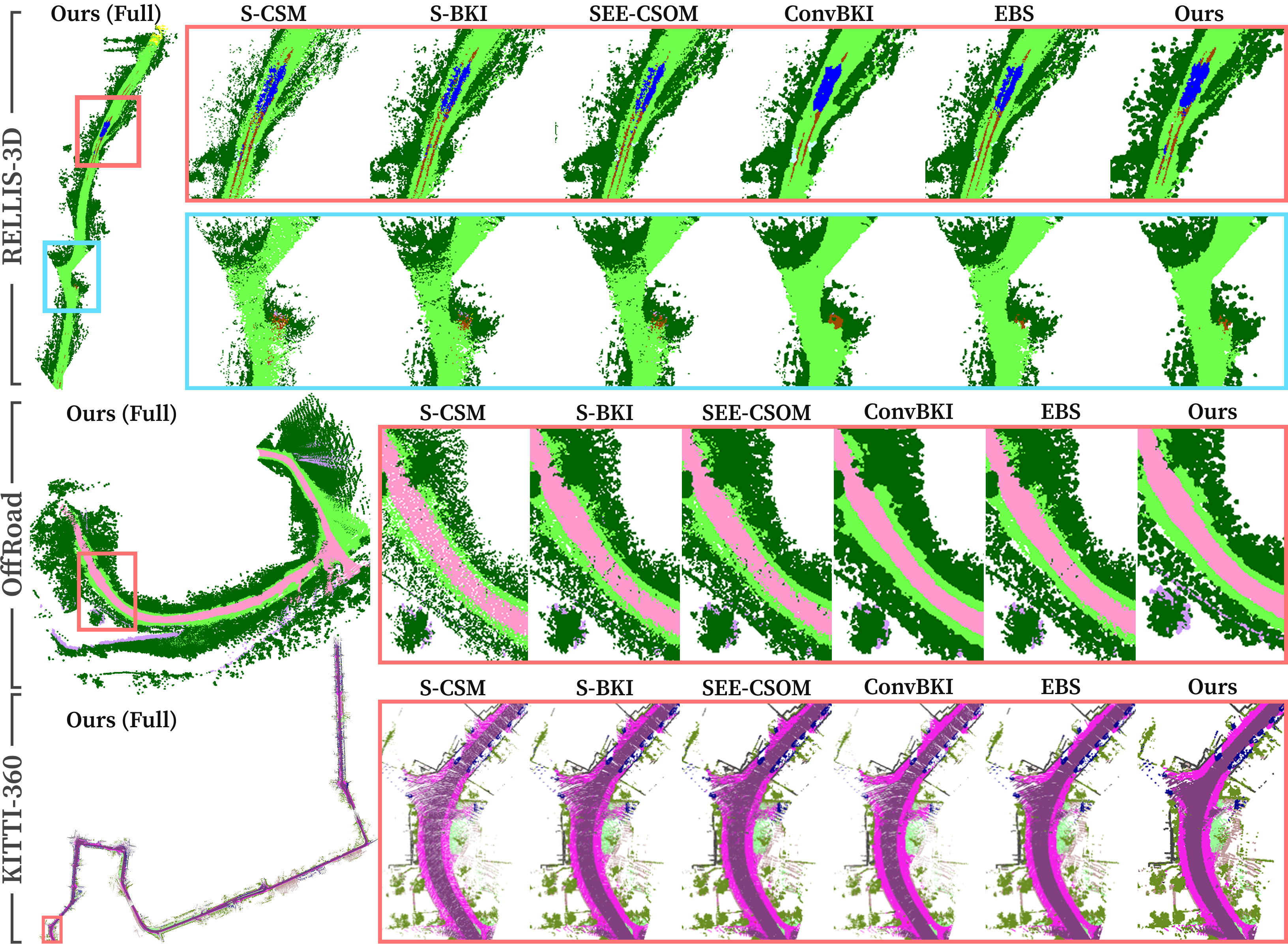}
\end{center}
\vspace{-0.1in}
\caption{Qualitative comparison of BEV semantic mapping results on \RELLIS, \OFFROAD, and \KITTI.}
\label{fig:aBEV}
\vspace{-0.1in}
\end{figure*}

\section{Supplementary Results}
\subsection{Quantitative Results}
We present more complete quantitative results corresponding to those reported in \Tref{tab:offroad_quant} and \Tref{tab:urban_quant}. Since our Gaussian primitives naturally support continuous evaluation, we include continuous evaluation results. Moreover, we report \ECE to provide a more comprehensive assessment of uncertainty calibration. The quantitative results are summarized in \Tref{atab:full_quant}.

\begin{table}[t!]
    \centering
    \renewcommand{\arraystretch}{1.0}
    \caption{Summary of semantic accuracy and uncertainty calibration on \RELLIS, \OFFROAD, and \KITTI. Higher is better for \mIoU and \Acc, lower is better for \BS and \ECE.}
    \label{atab:full_quant}
    \small{
        \resizebox{1.0\linewidth}{!}{
        \begin{tabular}{l | cccc | cccc | cccc}
        \toprule
        & \multicolumn{4}{c|}{\RELLIS} 
        & \multicolumn{4}{c|}{\OFFROAD} 
        & \multicolumn{4}{c}{\KITTI} \\
        \cmidrule(lr){2-5} \cmidrule(lr){6-9} \cmidrule(lr){10-13}
        \textbf{Method}
        & \rotatebox{90}{\mIoU [\%]} & \rotatebox{90}{\Acc [\%]} & \rotatebox{90}{\BS$\downarrow$ [\%]} & \rotatebox{90}{\ECE$\downarrow$ [\%]}
        & \rotatebox{90}{\mIoU [\%]} & \rotatebox{90}{\Acc [\%]} & \rotatebox{90}{\BS$\downarrow$ [\%]} & \rotatebox{90}{\ECE$\downarrow$ [\%]}
        & \rotatebox{90}{\mIoU [\%]} & \rotatebox{90}{\Acc [\%]} & \rotatebox{90}{\BS$\downarrow$ [\%]} & \rotatebox{90}{\ECE$\downarrow$ [\%]} \\
        \midrule \midrule
        \textit{S-CSM} &
            43.7 & 67.7 & 15.0 & 13.9 & 69.6 & 58.5 & 9.2 & 7.9 & 47.3 & 59.5 & 14.7 & 12.8 \\
        \textit{S-BKI} &
            43.8 & 78.9 & 16.3 & 12.4 & 69.2 & 78.5 & 14.4 & 12.6 & 47.1 & 74.9 & 17.4 & 10.8 \\
        \textit{SEE-CSOM} &
            44.6 & 80.0 & 14.1 & 11.6 & 68.8 & 80.3 & 8.7 & 4.8 & 46.9 & 76.1 & 13.4 & 8.2 \\
        \textit{ConvBKI} &
            44.3 & 80.7 & 15.2 & 14.7 & 71.1 & 84.3 & 8.4 & 7.3 & 48.3 & 68.0 & 13.5 & 11.4 \\
        \textit{EBS} &
            45.1 & 80.9 & 13.2 & 9.9 & \first{71.9} & 82.6 & \first{7.0} & 3.0 & 49.0 & 77.8 & 12.5 & \first{5.1} \\
        \textit{Ours} &
            \secnd{47.3} & \secnd{83.5} & \secnd{13.0} & \first{9.1} & 71.5 & \secnd{87.7} & 7.4 & \first{2.2} & \secnd{49.4} & \secnd{80.0} & \secnd{12.1} & 6.3 \\
        \textit{Ours (Cont.)} &
            \first{47.4} & \first{83.7} & \first{12.9} & \secnd{9.2} & \secnd{71.7} & \first{87.9} & \secnd{7.3} & \secnd{2.3} & \first{49.6} & \first{80.2} & \first{12.0} & \secnd{6.2} \\
        \bottomrule
        \end{tabular}
        }
    }
    \vspace{-0.2in}
\end{table}

\subsection{Semantic BEV Mapping}
\Fref{fig:aBEV} illustrates qualitative results of semantic BEV mapping experiments. On \RELLIS, our method effectively suppresses noisy vegetation predictions and yields cleaner boundaries along roads and traversable ground. On \OFFROAD, it preserves thin structures and produces sharp guardrail predictions that baselines often miss. On \KITTI, it reconstructs a more complete and consistent road surface even with sparse mapping input.

\subsection{Ablation Studies}
\begin{table}[t!]
\centering
\scriptsize
\renewcommand{\arraystretch}{1.0}
\caption{Ablation study evaluating the contribution of each component in our framework, alongside comparisons with ConvBKI variants.}
\label{tab:ablation_full}
\normalsize{
\resizebox{1.0\linewidth}{!}{%
        \begin{tabular}{l c ccc}
        \toprule
        \multicolumn{1}{c}{\textbf{Component(s)}} & \textbf{Section} & \mIoU & \Acc & \BS$\downarrow$ \\
        \midrule
        \multicolumn{1}{l}{S-BKI~\cite{51_S-BKI_gan2020}} & -- & 43.8 & 78.9 & 16.3 \\ 
        {\small \textit{+ Evidential BKI} (EBS~\cite{EBS_kim2024})}  &  \ref{Method:EDL} & 45.1 & 80.9 & 13.2 \\ 
        {\small \textit{+ FPS Clustering}} & \ref{Method:Init} & 45.3 & 82.8 & 14.3 \\
        {\small \textit{+ K\text{-}Means++ Clustering}} & \ref{Method:Init} & 45.5 & 83.1 & 14.4 \\
        {\small \textit{+ Anisotropic Gaussian Primitive}} & \ref{Method:Init} & 45.8 & 83.2 & 14.3 \\
   
        {\small \textit{+ Gaussian Refinement (Merging)}} & \ref{Method:Refine} & 45.9 & 83.2 & 13.9 \\
        {\small \textit{+ Gaussian Refinement (Pruning)}} & \ref{Method:Refine} & \secnd{47.3} & \secnd{83.5} & 13.4 \\ 
        {\small \textit{+ Uncertainty Decomposition (Ours)}}     & \ref{Appendix:UncDec} & \secnd{47.3} & \secnd{83.5} & \secnd{13.0} \\ 
        {\small \textit{+ Continuous Evaluation}}         & \ref{EXP:RobVer} & \first{47.4} & \first{83.7} & \first{12.9} \\
        \cmidrule(lr){1-5}
        \multicolumn{1}{l}{ConvBKI~\cite{47_ConvBKI2_wilson2023}} {\small w/o access to test data} & -- & 44.3 & 80.7 & 15.2 \\
        \multicolumn{1}{l}{ConvBKI~\cite{47_ConvBKI2_wilson2023}} {\small w/ access to test data} & -- & 45.4 & 81.7 & 14.7 \\
        \bottomrule
    \end{tabular}
    }
}
\vspace{-0.1in}
\end{table}
We conduct ablation studies on \RELLIS as summarized in \Tref{tab:ablation_full}. Starting from the S-BKI baseline, EBS addresses semantic uncertainty and demonstrates substantial improvements. The introduction of isotropic Gaussian primitives transitions the framework to primitive-based operations, allowing local observations to be aggregated into semantically coherent units. We also evaluate Farthest Point Sampling~(FPS), but K-Means++ yields more balanced clusters and further stabilizes performance. Extending primitives to anisotropic Gaussians enables geometry-aligned kernels, mitigating spatial uncertainty. Local merging improves mapping stability and computational efficiency without sacrificing accuracy, while adaptive pruning significantly enhances semantic performance. Finally, uncertainty decomposition in \eqref{E2BKI_unc} improves posterior calibration, resulting in more reliable uncertainty estimates, as shown in \Fref{fig:avarmap}.

To make the effect of uncertainty decomposition clear, \Fref{fig:avarmapone} presents single-frame semantic and uncertainty maps for the same scene as \Fref{fig:avarmap}. The baseline uncertainty map is dominated by local evidential sparsity, visible as periodic patterns, whereas the decomposed maps separate sparsity from semantic ambiguity and show better calibration.

We further validate our design choices by comparing ConvBKI with and without access to test data. While its performance improves by using test data during training, it is outperformed by our method even with only isotropic Gaussians, highlighting its limitations in handling intra-class variation due to fixed class-wise kernels. In contrast, our approach generalizes effectively through geometry-aligned Gaussian modeling without relying on ground truth labels.

\subsection{Sensitivity Analysis}
We conduct a sensitivity analysis to evaluate the robustness of E2-BKI with respect to its hyperparameters. E2-BKI involves the following hyperparameters: $\alpha_0^c$, $\ell$, $\beta$, $\tilde{u}$, $\epsilon$, $d_L$, $d_S$ (\Tref{table:hyperparams}), and $\tau$~\eqref{aGauss_tau}. Following prior work~\cite{51_S-BKI_gan2020,EBS_kim2024}, we fix the Dirichlet prior $\alpha_0^c = 0.001$ and set the kernel length-scale $\ell$ equal to the voxel resolution ($0.2\mathrm{m}$).

\noindent\textbf{Uncertainty Threshold}
The dynamic uncertainty threshold $\tilde{u}$ controls whether an incoming semantic observation is integrated into the map based on its predicted uncertainty. This mechanism is adopted from~\cite{EBS_kim2024} and is intended to suppress unreliable updates without introducing a fixed, dataset-specific cutoff. We analyze the effect of $\tilde{u}$ separately by explicitly comparing the proposed dynamic thresholding scheme against several static thresholds.

\begin{figure*}[t]
\centering
\includegraphics[width=1.0\linewidth]{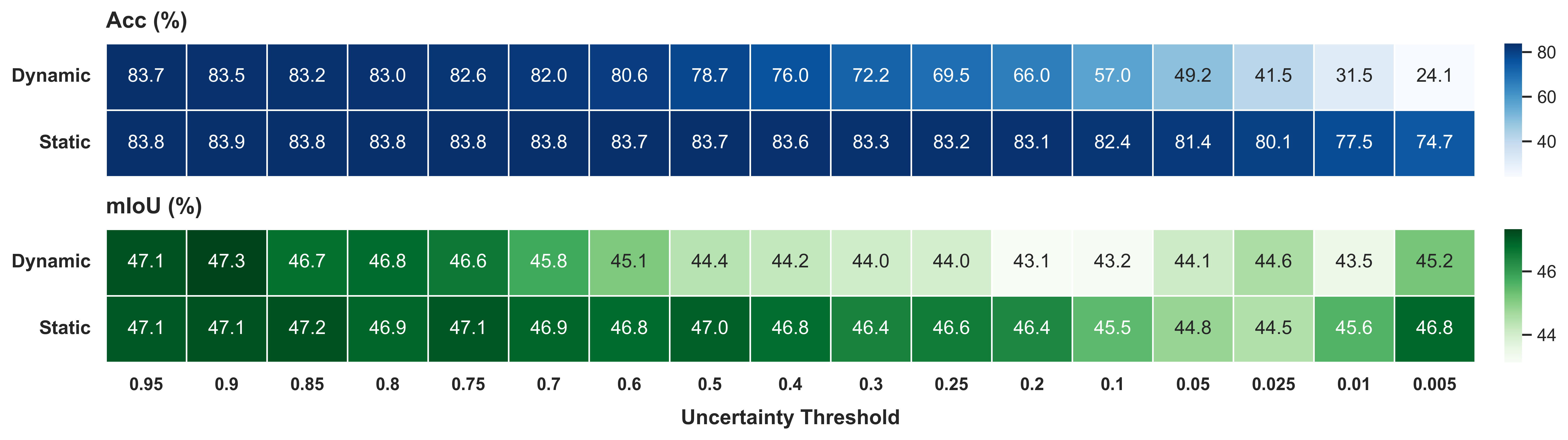}
\vspace{-0.2in}
\caption{Effect of the dynamic and static uncertainty threshold on \Acc and \mIoU on \RELLIS.}
\vspace{-0.1in}
\label{fig:aUncExp}
\end{figure*}

Concretely, we vary the uncertainty threshold over the range $[0.05, 0.95]$ and evaluate both strategies. In the dynamic scheme, the hyperparameter $\tilde{u}$ specifies the proportion of the most uncertain primitives to be removed in each frame. For clarity, the horizontal axis in \Fref{fig:aUncExp} is parameterized as $\texttt{Uncertainty Threshold} = 1 - \tilde{u}$. For example, a threshold value of $0.9$ corresponds to removing the top $10\%$ most uncertain primitives per frame in the dynamic scheme, whereas the static scheme removes all primitives whose uncertainty exceeds $0.9$ globally.

The resulting \Acc and \mIoU are summarized in \Fref{fig:aUncExp}. As the threshold decreases, fewer primitives are retained, leading to degraded \Acc for both schemes. For \mIoU, the dynamic strategy achieves its best performance at a threshold of $0.9$ ($\tilde{u} = 10\%$) and consistently outperforms any static threshold. This is because dynamic thresholding adapts to frame-wise variations in the uncertainty distribution, whereas a global static threshold is difficult to tune and may remove too few or too many primitives depending on the scene.

Based on these observations, we adopt the dynamic threshold with $\tilde{u} = 10\%$ as our default. This setting provides stable performance and generalization across datasets without additional tuning, in contrast to EBS~\cite{EBS_kim2024} which relies on dataset-specific values ($\tilde{u} = 10\%$ on \RELLIS and $\tilde{u} = 30\%$ on \OFFROAD). We note that when focusing on a single dataset or a specific downstream task, a carefully tuned static threshold may be beneficial. A hybrid strategy that combines dynamic thresholding with an additional static cutoff is a promising direction for future work.

\noindent\textbf{Additional Hyperparameters}
\begin{figure*}[t]
\centering
\includegraphics[width=1.0\linewidth]{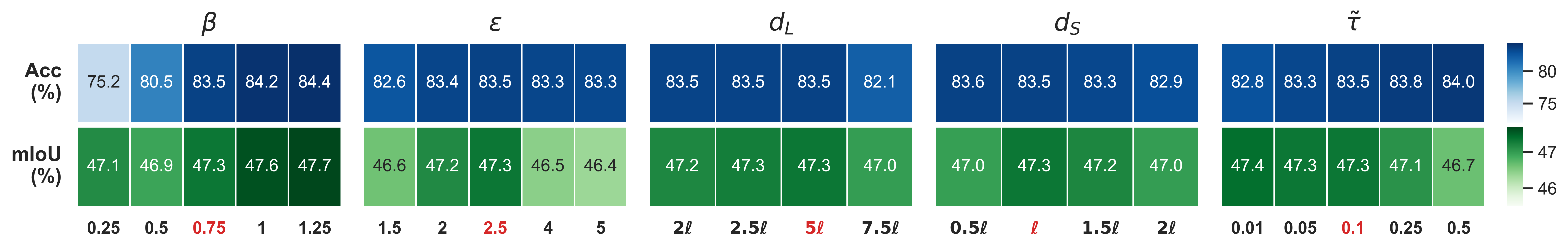}
\caption{Sensitivity analysis of the five hyperparameters used in E2-BKI on \RELLIS. For each hyperparameter, only that variable is varied while all others are fixed to the default configuration ($\beta = 0.75,\ \epsilon = 2.5,\ d_L = 5.0 \ell,\ d_S = \ell,\ \tilde \tau = 0.1$). The definition of $\tilde{\tau}$ is provided in \eqref{aGauss_tau}.}
\label{fig:aParamExp}
\vspace{-0.1in}
\end{figure*}
\begin{figure*}[t]
\centering
\includegraphics[width=1.0\linewidth]{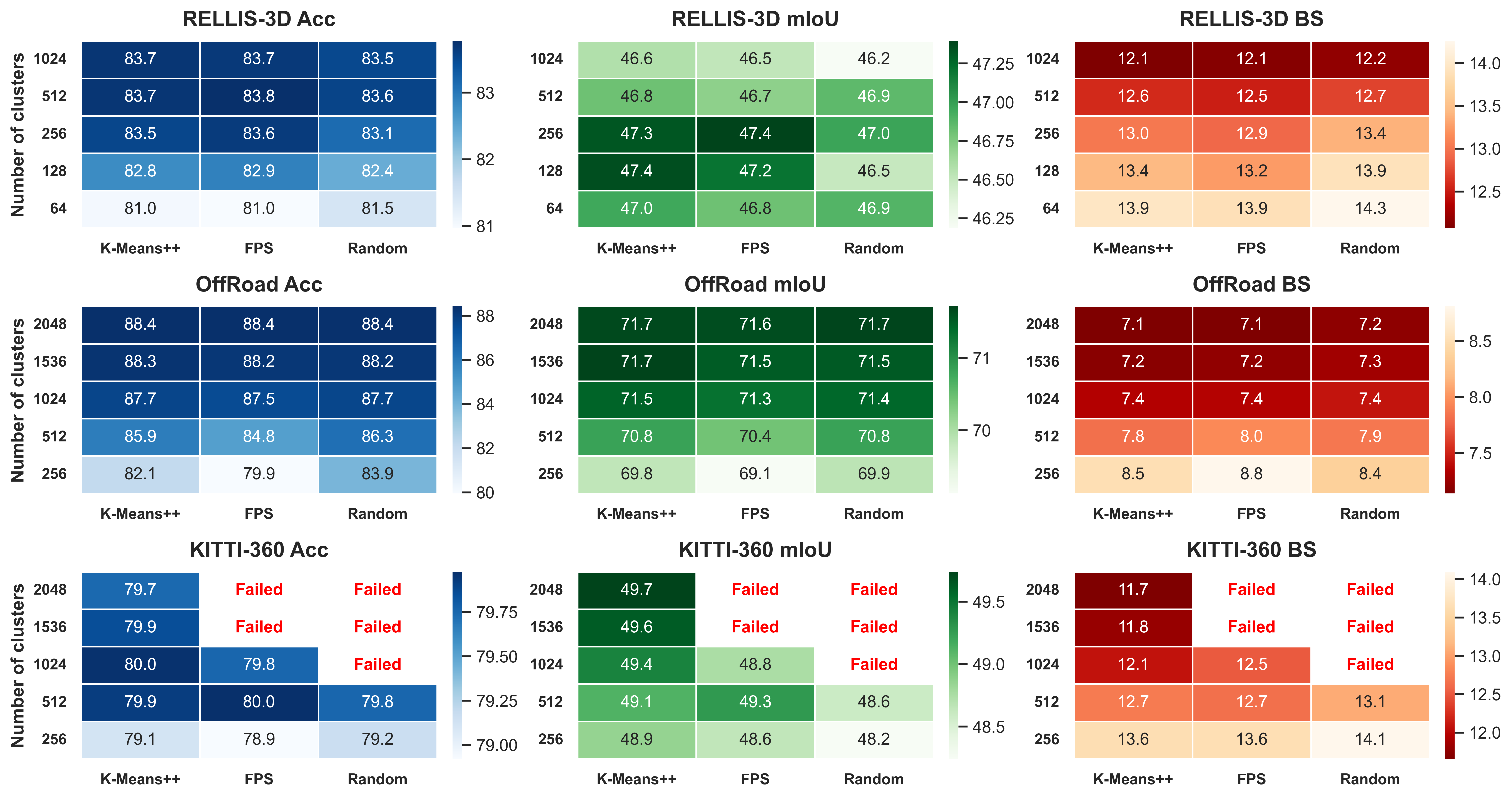}
\caption{Sensitivity to the clustering strategy (K-Means++, FPS, and Random sampling) and the number of clusters.}
\label{fig:aInitExp}
\vspace{-0.2in}
\end{figure*}
\Fref{fig:aParamExp} reports the effect of varying each hyperparameter while fixing others to the default values ($\beta = 0.75, \epsilon = 2.5, d_L = 5.0 \ell, d_S = \ell, \tilde{\tau} = 0.1$) on \RELLIS. Both \Acc and \mIoU remain highly stable across a wide range of values. Notably, even with hyperparameter variations, our method consistently outperforms the best baseline results (\mIoU$ = 45.1$ and \Acc$ = 80.9$), indicating that the performance gain stems from the proposed Gaussian-primitive formulation rather than hyperparameter tuning. While larger $\beta$ improves accuracy, it increases inference latency. Thus, we select $\beta = 0.75$ as a practical trade-off between accuracy and runtime.

A grid search over $2{,}000$ configurations revealed that $287$ settings ($14\%$) surpassed our default performance, with the best configuration reaching \Acc$ = 84.3$ and \mIoU$ = 48.3$. However, we deliberately employ the fixed default configuration throughout our experiments. This demonstrates that E2-BKI is robust to hyperparameter choices and generalizes well without requiring exhaustive, dataset-specific tuning.

\subsection{Clustering Strategy Analysis}
While E2-BKI is theoretically agnostic to the clustering strategy used to form local point groups, the initialization quality can impact geometric fidelity and numerical stability. To examine the effect of different initialization schemes, we evaluate three approaches: K-Means++, Farthest Point Sampling~(FPS), and Random sampling. \Fref{fig:aInitExp} summarizes the quantitative results (\Acc, \mIoU, and \BS) on \RELLIS, \OFFROAD, and \KITTI. While performance trends are consistent across strategies, K-Means++ demonstrates superior stability. FPS and Random sampling occasionally fail at high cluster counts (marked as `Failed' in \Fref{fig:aInitExp}) due to insufficient points for valid covariance estimation, whereas K-Means++ remains stable.

\begin{figure}[t]
\centering
\includegraphics[width=1.0\linewidth]{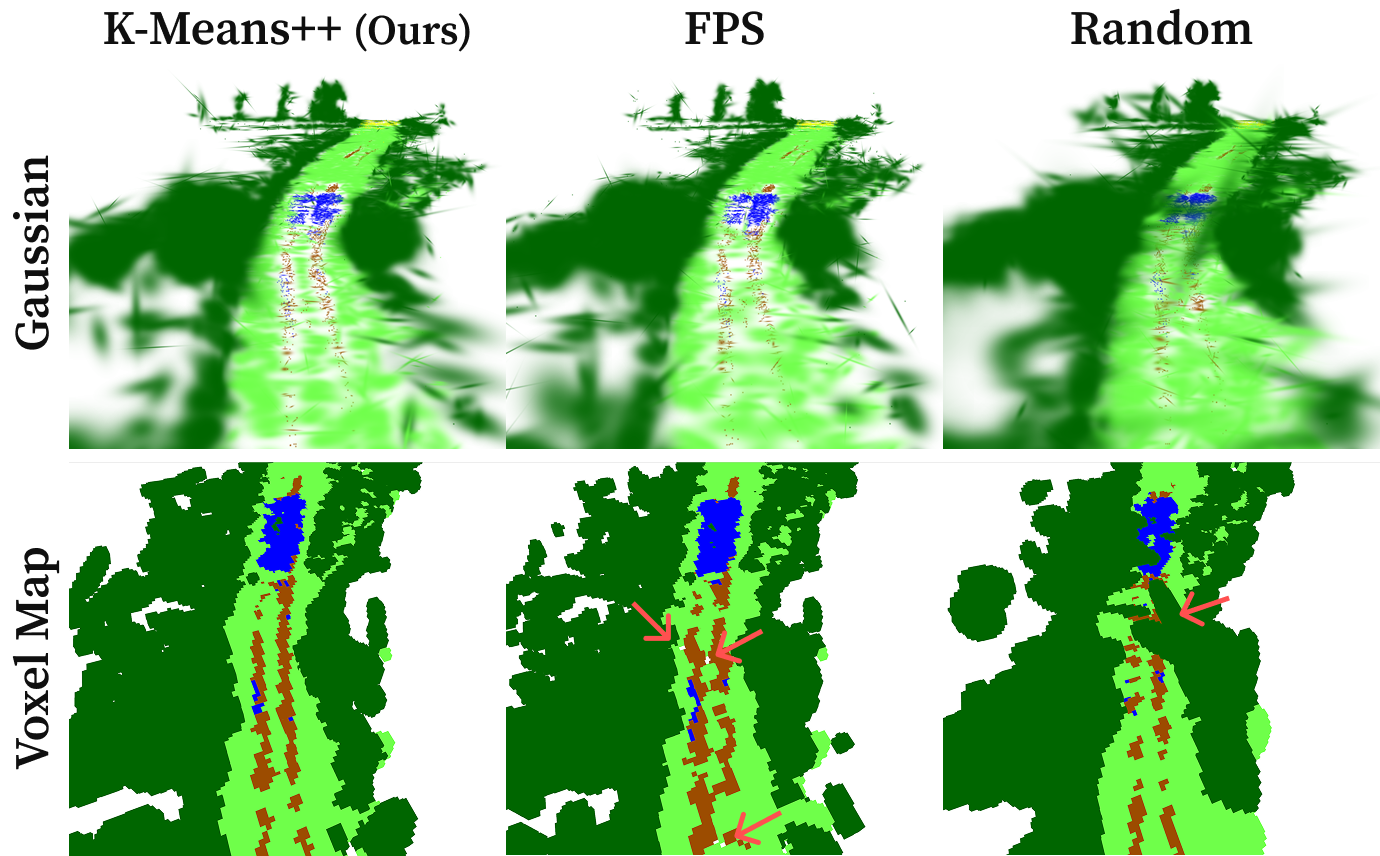}
\caption{Qualitative comparison of clustering strategies on \RELLIS using only $64$ clusters. The top row shows Gaussian primitives' renderings, and the bottom row shows voxelized semantic maps. Red arrows highlight representative artifacts for FPS and Random sampling.}
\label{fig:aInitExpQual}\vspace{-0.1in}
\end{figure}

As visualized in \Fref{fig:aInitExpQual}, the qualitative comparison on \RELLIS with only $64$ clusters reveals that K-Means++ preserves fine geometric details, whereas FPS and Random sampling exhibit artifacts such as holes or over-inflation. Based on these observations, we select K-Means++ as our default initialization strategy due to its stability. Regarding the cluster count, we employ $256$ clusters for \RELLIS, while scaling up to $1{,}024$ clusters for \OFFROAD and \KITTI to sufficiently capture the richer geometric details present in these environments.

\subsection{Extended Runtime Analysis}
We present an extended runtime analysis on \RELLIS in \Tref{tab:aRuntime}. All methods were evaluated on a laptop equipped with an Intel i7-12700H CPU and an NVIDIA RTX 3080 Ti Laptop GPU. The mapping speeds are given as the mean and standard deviation over five runs. While ConvBKI~\cite{47_ConvBKI2_wilson2023} attains the fastest runtime via GPU acceleration, it necessitates a substantial training overhead ($1759.5 \pm 128.3$ seconds on \RELLIS). However, E2-BKI is training-free and could be further accelerated by adopting GPU acceleration.

\begin{table}[t]
    \footnotesize
    \centering
    \renewcommand{\arraystretch}{1.2}
    \caption{Mapping speed (Hz) comparison.}
    \resizebox{\columnwidth}{!}{%
    \begin{tabular}{c|c|c|c|c|c}
        \hline
        S-CSM & S-BKI & SEE-CSOM & ConvBKI & EBS & Ours \\
        \hline
        3.44 $\pm$ 0.01 & 3.15 $\pm$ 0.04 & 3.00 $\pm$ 0.01 & \textbf{8.40 $\pm$ 0.21} & 1.65 $\pm$ 0.01 & 5.39 $\pm$ 0.11 \\
        \hline
    \end{tabular}%
    }
    \label{tab:aRuntime}
\vspace{-0.15in}
\end{table}

\noindent\textbf{Runtime Scalability}
\begin{figure}[t]
\centering
\includegraphics[width=1.0\linewidth]{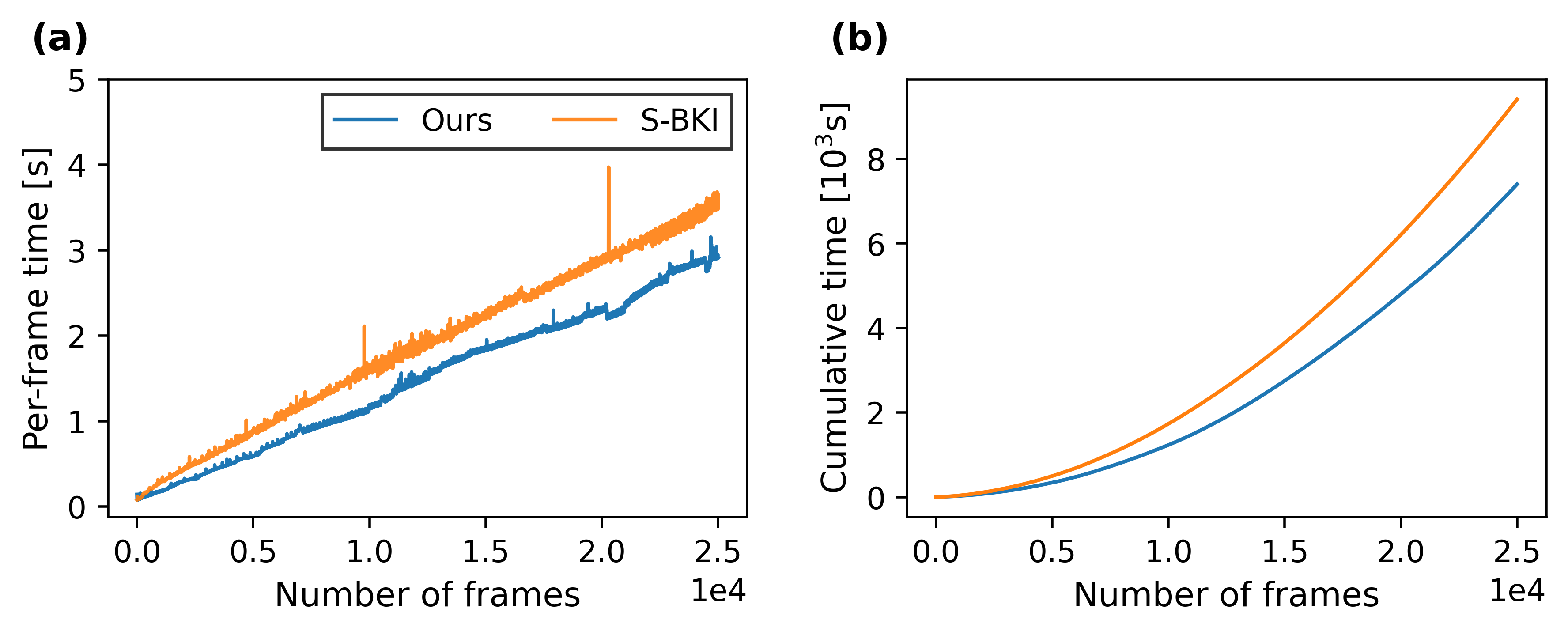}
\vspace{-0.1in}
\caption{\textbf{(a)} Per-frame and \textbf{(b)} cumulative runtime over long-term sequences.}
\label{fig:aTimeScale}
\begin{subfigure}{0pt}\phantomsubcaption\label{fig:aTimeScale:a}\end{subfigure}
\begin{subfigure}{0pt}\phantomsubcaption\label{fig:aTimeScale:b}\end{subfigure}
\vspace{-0.1in}
\end{figure} 
To evaluate the computational efficiency of E2-BKI in large-scale scenarios, we analyze its scalability relative to S-BKI~\cite{51_S-BKI_gan2020} using a synthetic sequence of straight-line motion spanning $25{,}000$ frames. This setup simulates a significantly longer operation compared to the \RELLIS sequences used in the main experiments (avg. $1{,}020$ frames), thereby assessing the long-term behavior.

The results are presented in \Fref{fig:aTimeScale}. As the trajectory continuously explores new environments, the number of Gaussian primitives $J$ grows linearly with the number of frames $F$. Since the current implementation manages primitives in a global set, the update cost scales with $J$, resulting in the observed quadratic growth ($\mathcal{O}(F^2)$) in cumulative runtime (\Fref{fig:aTimeScale:b}). Despite this, E2-BKI consistently maintains a lower computational cost than S-BKI. It is important to note that this straight-line motion represents a worst-case scenario for map expansion; in realistic deployments with revisited regions, merging and pruning mechanisms would effectively limit primitive growth. Furthermore, this quadratic scaling is an implementation artifact rather than a fundamental limitation. By introducing block-wise partitioning to bound the number of active primitives during updates, the system can achieve linear scalability $\mathcal{O}(F)$.

\noindent\textbf{Impact of Voxel Resolution and Input Density}
\begin{figure}[t]
\centering
\includegraphics[width=1.0\linewidth]{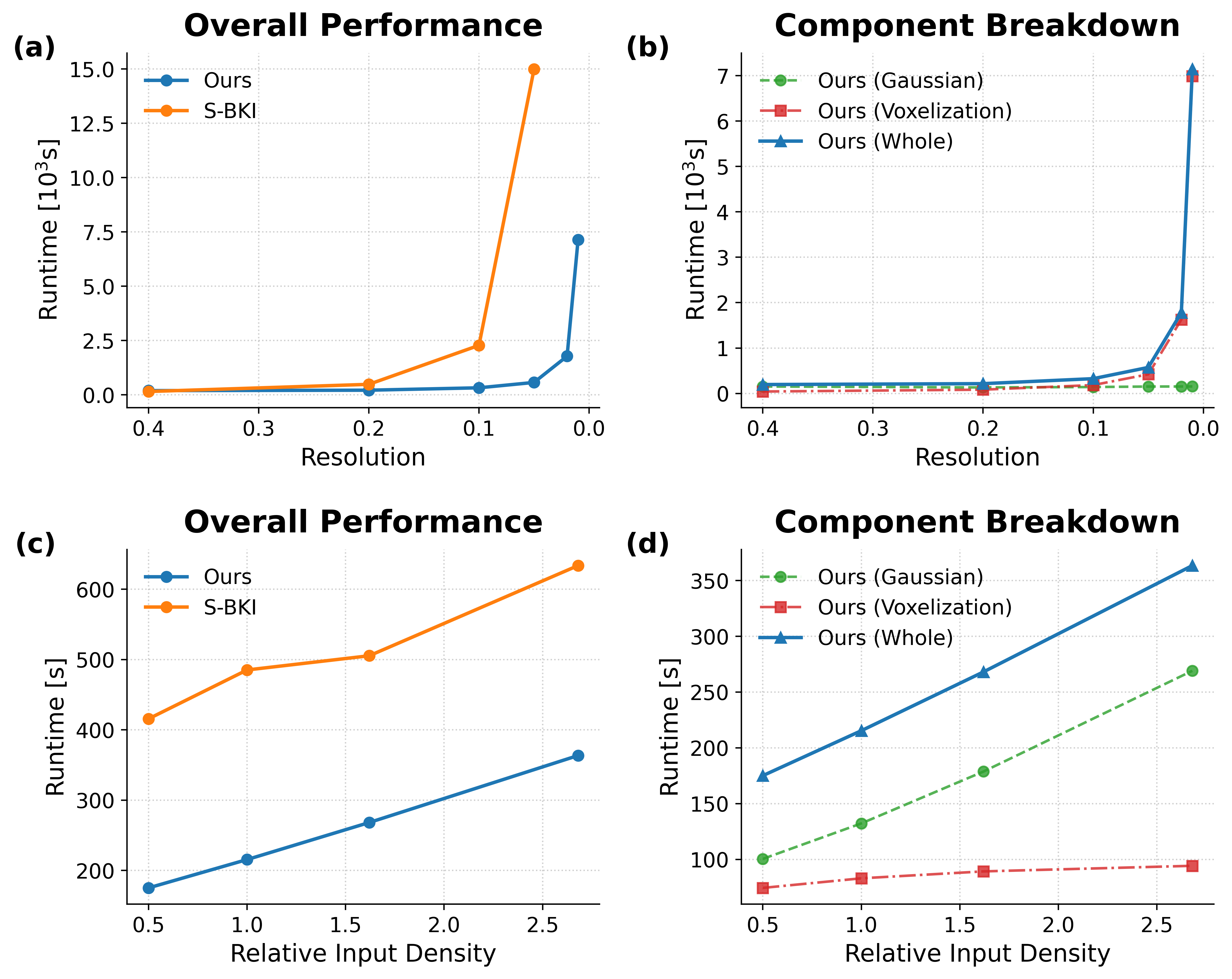}
\vspace{-0.15in}
\caption{\textbf{Runtime scalability of E2-BKI.} The left \textbf{(a, c)} compare the total runtime of E2-BKI and S-BKI across voxel resolutions and relative input densities. The right \textbf{(b, d)} provides a component-wise breakdown of E2-BKI, separating Gaussian construction from voxelization.}
\label{fig:aResExp}
\begin{subfigure}{0pt}\phantomsubcaption\label{fig:aResExp:a}\end{subfigure}
\begin{subfigure}{0pt}\phantomsubcaption\label{fig:aResExp:b}\end{subfigure}
\begin{subfigure}{0pt}\phantomsubcaption\label{fig:aResExp:c}\end{subfigure}
\begin{subfigure}{0pt}\phantomsubcaption\label{fig:aResExp:d}\end{subfigure}
\vspace{-0.1in}
\end{figure}
We evaluate the computational scalability of E2-BKI by varying the voxel resolution from $0.4\mathrm{m}$ down to $0.01\mathrm{m}$ and adjusting the input point density. As illustrated in \Fref{fig:aResExp:a}, S-BKI becomes prohibitively slow at fine resolutions ($<0.05\mathrm{m}$), whereas E2-BKI remains tractable even at $0.01\mathrm{m}$. Notably, our method at $0.02\mathrm{m}$ operates faster than S-BKI at $0.1\mathrm{m}$, demonstrating superior efficiency. Furthermore, \Fref{fig:aResExp:c} confirms that E2-BKI scales linearly with input density. While S-BKI exhibits a steep increase in runtime, our framework grows at a moderate rate, reflecting natural data volume growth without significant algorithmic overhead.

The component-wise breakdown (\Fref{fig:aResExp:b}) reveals that the Gaussian construction time is invariant to voxel resolution, confirming that our primitive generation is decoupled from the output grid. The exponential increase in runtime at extreme resolutions ($0.01\mathrm{m}$) is driven solely by the voxelization process, stemming from the inherent memory overhead of dense grids rather than the proposed Gaussian formulation. Regarding input density, \Fref{fig:aResExp:d} shows that Gaussian construction grows proportionally to the number of points, while voxelization remains relatively stable. This indicates that E2-BKI efficiently handles denser sensor streams.